\documentclass[10pt,journal,compsoc]{IEEEtran}
\usepackage{cite}
\usepackage{verbatim}
\usepackage{amsmath,amssymb,amsfonts}
\usepackage{algorithmic}
\usepackage{graphicx}
\usepackage{textcomp}
\usepackage{xcolor}
\usepackage{cite}
\usepackage{caption} 
\usepackage{multicol}
\usepackage[ruled,linesnumbered]{algorithm2e}
\usepackage[normalem]{ulem}

\def\ournotes{1}
\newcommand{\del}[1]{\textbf{\textcolor{red}{\if\ournotes1 \sout{#1} \fi}}}
\newcommand{\todo}[1]{\textbf{\textcolor{red}{\if\ournotes1 TODO: #1 \fi}}}
\newcommand{\eli}[1]{\textbf{\textcolor{cyan}{\if\ournotes1 ETB: #1 \fi}}}
\newcommand{\jiahao}[1]{\textbf{\textcolor{magenta}{\if\ournotes1 JD: #1 \fi}}}
\newcommand{\fix}[2]{\textbf{\textcolor{blue}{\if\ournotes1 \sout{#1} {#2} \fi}}}

\ifCLASSINFOpdf
\else
\fi
\begin{document}

%
\title{SSDBCODI: Semi-Supervised Density-Based Clustering with Outliers Detection Integrated}
%
%
%
%

\author{Jiahao~Deng~\IEEEmembership{}
        and~ Eli~T.~Brown,~\IEEEmembership{}
\IEEEcompsocitemizethanks{\IEEEcompsocthanksitem J. Deng and E. Brown are with the College of Computing and Digital Media, DePaul University, Chicago,
IL, 60604.\protect\\}}

%
%

\markboth{}%
{SSDBCODI: Semi-Supervised Density-Based Clustering with Outliers Detection Integrated}
%



\IEEEtitleabstractindextext{%
\begin{abstract}
Clustering analysis is one of the key tasks in machine learning. Traditionally, clustering has been an independent task, separate from outlier detection. Due to the fact that the performance of clustering can be significantly eroded by outliers, a small number of algorithms try to incorporate outlier detection in the process of clustering. However, most of those algorithms are based on unsupervised  partition-based algorithms such as k-means. Given the nature of those algorithms, they often fail to deal with clusters of complex, non-convex shapes. To tackle this challenge, we have proposed SSDBCODI, a semi-supervised density-based algorithm.
SSDBCODI combines the advantage of density-based algorithms, which are capable of dealing with clusters of complex shapes, with the semi-supervised element, which offers flexibility to adjust the clustering results based on a few user labels. We also merge an outlier detection component with the clustering process. Potential outliers are detected based on three scores generated during the process: (1) reachability-score, which measures how density-reachable a point is to a labeled normal object, (2) local-density-score, which measures the neighboring density of data objects, and (3) similarity-score, which measures the closeness of a point to its nearest labeled outliers. Then in the following step, instance weights are generated for each data instance based on those three scores before being used to train a classifier for further clustering and outlier detection. Since our algorithm aims to combine both outlier detection and clustering, to enhance the understanding of the proposed algorithm, for our evaluation, we have run our proposed algorithm against some of the state-of-art approaches on multiple datasets and separately listed the results of outlier detection apart from clustering. Our results indicate that our algorithm can achieve superior results with a small percentage of labels.
\end{abstract}

}

\maketitle

\IEEEdisplaynontitleabstractindextext

%
\IEEEpeerreviewmaketitle


\section{Introduction} \label{Intro}
Clustering as a sub-field in machine learning studies ways to group data objects and has wide applications in many areas such as finance \cite{hernandez2020fuzzy}, energy\cite{li2021monitoring,younis2004heed}, medicine \cite{veloso2014clustering} and mechanical engineering\cite{rose1990statistical}.

Normally, clustering tasks assume every instance belongs to one designated cluster group. However, this assumption is often misleading due to the fact that outliers widely exist in real-world data. This poses great challenges to many partition-based algorithms given the fact that incorrect inclusion of certain outliers into a cluster may shift its center. Traditionally, outlier detection has been a separate task from clustering analysis, but some researchers have incorporated outlier detection into clustering analysis \cite{hautamaki2005improving, chen2020robust, liu2019clustering}. These approaches have achieved good results, but they face one major limitation. This limitation revolves around the fact that most of those approaches are partition-based.  In fact, according to Rose et al.~\cite{rose1990statistical} and Bhargav et al.~\cite{bhargav2016review}, partition-based algorithms contribute one of the largest shares of the clustering techniques. This type of algorithm assumes distinct clusters form convex shapes and that objects inside the same cluster share a higher level of similarity than data of different clusters. For example, k-means, one of the earliest adopted techniques in the field but also in common usage today, focuses on partitioning the data into different groups based on constantly updating the centers of each cluster until convergence. However, as Xu et al.\cite{rose1990statistical,bhargav2016review} illustrate, partition-based algorithms often assume clusters to be convex and outliers are simply points distant from the centroids of each cluster. Therefore, partition-based algorithms are often challenged by clusters in non-convex or complex shapes as well as data of varying densities, impacting the quality of the outlier detection process.

To tackle the limitation of the partition-based algorithms, we rely on density-based algorithms. The idea of the density-based algorithms is that they group adjacent data objects of the same cluster based on their neighboring density. For example, DBSCAN is one of the earliest algorithms in this category. This algorithm starts with some root objects and iteratively searches for the neighboring data that are density-connected to the root objects. There are several advantages to this type of algorithm: first, since this algorithm only groups data located in high-density areas, it will ignore common types of noise points. Secondly, it can capture clusters of irregular shapes because the cluster was expanded by evaluating every single point instead of applying a global partition.

Traditionally, clustering is an unsupervised task, not requiring any labels. 
However, as mentioned earlier, in reality, the data sometimes contain a few labels for certain objects or other information such as must-link or cannot-link constraints. 
That information could provide useful and subtle context for both cluster analysis and outlier detection \cite{gao2006semi,bair2013semi}. Benefiting from this idea, semi-supervised clustering is a type of method that takes advantage of those labeled objects while still considering and clustering the unlabeled objects. To incorporate the semi-supervised learning component into the density-based clustering, SSDBSCAN was proposed by Lelis et al.~\cite{lelis2009semi}. This algorithm starts with each of the labeled data objects as the root points and expands from the root points to the neighboured unlabeled points until it reaches another labeled point $p$ that belongs to a different cluster. During the expansion process, the maximum density threshold is computed accordingly for each cluster to ensure that points of different labels are not grouped together into the same cluster. The advantage of this algorithm is that it does not assume a global density threshold and ensures each cluster includes as many data points as possible while respecting the constraints of labeled data points (data of different labels should not end up in the same cluster). Although experiments indicate that this algorithm shows robustness in handling clusters of various densities compared to the regular DBSCAN, the existence of certain outliers will still reduce its accuracy given the fact that after clustering, certain data points will remain unclustered and it lacks an effective measure to explicitly detect and separate the actual outliers from the normal points.

Inspired by this, we propose SSDBCODI, a semi-supervised density-based algorithm, mainly based on three assumptions in outlier detection and other density-based clustering algorithms. 
First, outliers are typically less density-reachable to clusters than other points.
Second, outliers are located in the less dense regions as they are often separated from the majority of data \cite{zhang2018anomaly}.
Finally, outliers are relatively closer to each other than to the other points \cite{chandola2009anomaly}.
Utilizing those assumptions, SSDBCODI works in three steps. First, we run a modified version of SSDBSCAN to identify non-anomaly points and record three properties: \textit{Reachability-Score}, which measures how densely reachable a point is to its nearest non-anomaly-labeled point,  \textit{Local-Density-Score}, which measures how isolated a point is based on its local density, and \textit{Similarity-Score}, which measures how close a point is to the nearest user-labeled outlier. Second, we use the weighted average of those three scores to detect outliers. The use of those scores offers the advantage of not only detecting potential outliers but also quantifying their outlierness. Lastly, we use a classification model to predict cluster membership and outliers.
We use the \textit{Reachability-Scores} as instance weights for the points identified as non-anomalies in step 1.
For outliers identified by step 2, the instance weights are a weighted average of the three scores from step 1.

Our proposed algorithm is the first density-based algorithm that incorporates both semi-supervised learning and explicit outlier detection into clustering to simultaneously cluster data and detect outliers. Other density-based algorithms such as DBSCAN and SSDBSCAN either lack the semi-supervised component or mechanisms to explicitly quantify the outlierness. Through extensive experiments, we demonstrate the effectiveness of our approach on different datasets.

The paper is organized as follows: in Section \ref{related}, we introduce the related work and contributions to the field. We explore the details of \textit{SSDSCAN} as it lays the foundation for our work in Section \ref{pre}.  In Section \ref{method}, we explain the new clustering and outlier detection algorithm we propose. Finally, Section \ref{eval} illustrates the evaluation results of our new approach compared to other baseline algorithms.
\section{Related Work} \label{related}
In this section, we discuss existing work that relates to this manuscript from multiple areas: outlier detection in general, then density-based clustering methods and semi-supervised clustering.  Finally we discuss recent work in clustering with outlier removal.

\subsection{Outlier Detection}
Outlier detection is one of the most studied areas in machine learning and is often referred as anomaly detection depending on the usage scenario. Traditionally outlier detection is an unsupervised learning task, due to the fact that outliers only account for a small percentage of the whole data. According to Chandola et al.~\cite{chandola2009anomaly}, some of the most used outlier detection methods can be categorized as follows:
\begin{itemize}
    \item Clustering-based algorithms: these types of algorithms are mainly based on the assumptions that normal instances are usually closer to (or share higher similarity to) the centroids or mean values of normal cases inside clusters compared to outliers. Typical examples include Kannan et al.~\cite{kannan2017outlier}, which uses matrix factorization for text anomaly detection, Alvarez's \cite{marcos2013clustering} multi-view outlier detection approach, and Pamula et al.~\cite{pamula2011outlier}, a k-means-based approach.
    \item Nearest Neighbor Based algorithms, which utilize the assumption that outliers generally occur in sparse regions while normal instances occur in dense areas. Local outlier factor (LOF) \cite{breunig2000lof}, is a state-of-art method in this category.
    \item Statistically Based algorithms are the ones utilizing various statistical properties including entropy \cite{yuan2021fuzzy}, similarity between cases \cite{kriegel2008angle}, deviation from normal instances \cite{leys2013detecting} and correlation \cite{papadimitriou2003loci}.
    \item Model-Based algorithms refer to methods that compare potential outliers to a constructed model of the data. They have gained popularity lately due to their flexibility.  Well-known approaches include One-Class SVM (OSVM) \cite{zhang2007one}, isolation forest \cite{liu2008isolation} and neural-network-based approaches  \cite{zenati2018efficient,li2021monitoring,an2015variational}.
\end{itemize}
Generally, these algorithms are unsupervised and rely on a single core assumption. This negatively affects their flexibility to adjust for diverse and complex situations. Inspired by this, our algorithm provides a semi-supervised mechanism 
to automatically adapt to few user labels.

\subsection{Density-Based Clustering}
As discussed in Section \ref{Intro}, partition-based approaches generally assume clusters form convex shapes. However, this assumption causes trouble since useful data groups are frequently shaped in complex, non-convex ways. Density-based algorithms can serve as a solution because they instead assume data points of the same clusters are more densely connected to each other than they are to points of different clusters.
DBSCAN \cite{ester1996density} is one of the earliest density-based algorithms. The basic idea of this algorithm is to link data points in dense regions while leaving out noise points located in sparse regions, using a density threshold $\epsilon$ as a key hyper-parameter.
Many later algorithms are largely developed around DBSCAN. For example, OPTICS \cite{ankerst1999optics} improves upon it by utilizing the augmented ordering of a dataset. HDBSCAN \cite{campello2013density} combines hierarchical clustering and DBSCAN. Despite many advantages of these density algorithms, it is notoriously hard to set the right density threshold, e.g.~$\epsilon$ for DBSCAN. Given this challenge, SSDBSCAN takes advantage of user labels to actively adjust the density threshold for each cluster. 
Our algorithm uses this concept with an adapted threshold mechanism that fits our goal of integrating outlier detection.

\subsection{Semi-Supervised Clustering}
Semi-Supervised clustering rests on the idea that users can often bring  useful insight to the process of clustering analysis. Most of those semi-supervised clustering algorithms are built upon the existing clustering approaches. For example, HISSCLU \cite{bohm2008hissclu} modifies the traditional OPTICS algorithm by starting the expansion simultaneously from all labeled objects.  Wagstaff et al.~\cite{wagstaff2001constrained} propose a k-means-based algorithm incorporating must-link and cannot-link constraints. Several techniques since have built upon that success  \cite{bilenko2004integrating,davidson2006measuring}. Also as mentioned in Section \ref{Intro}, based on the DBSCAN algorithm, SSDBSCAN introduced by Lelis et al.\cite{lelis2009semi}, similar to HISSCLU, starts its expansion from each of the labeled objects. However, this expansion terminates when it encounters another labeled object with a different category. It does not depend on a global $\epsilon$ and will adjust the shape and size of each cluster based on the labels. Given this flexibility, our own approach is largely based on this and thus we will explore it in more detail in Section \ref{Pre-algo}.

\subsection{Robust Clustering with Outlier Detection}
As explained in Section \ref{Intro}, the existence of outliers will often erode the performance of clustering algorithms. Because of this, many algorithms incorporate mechanisms to attach more importance to data points with a higher degree of certainty of normality, while ignoring points likely to be outliers. For example, many density-based algorithms \cite{davis2007information,yi2012semi,ester1996density,nagaraja2020similarity} including DBSCAN will only run clustering on points located in dense regions while the remaining points are vaguely classified as noise. Although those algorithms show robustness against outliers, they do not explicitly detect outliers during the clustering process. Inspired by this, researchers have tried combining outlier detection into some of the unsupervised clustering techniques. For example, Hautamäki et al.~\cite{hautamaki2005improving} propose a k-means-based approach, ORC, that identifies and removes outliers based on their distances to the centroids of each cluster. Jiang et al.~\cite{jiang2001two} partitions the data space using a minimum spanning tree (MST) and then identifies the outliers based on sparsity. Liu et al.~propose COR, a k-means-based method that projects the original data onto a partition space to identify and remove the outliers through the compactness of each cluster. Additional approaches  focus on the modification to the objective function of the existing k-means clustering to specifically separate outliers from normal instances
\cite{chen2020robust,gan2017k}.

Inspired by the previous works about joint clustering with outlier removal, our work aims to address two issues of the existing approaches.  First, most of the existing approaches rely on partition-based methods. However, as mentioned in Section \ref{Intro}, they often assume the clusters to be convex, resulting in poor performance in  
some non-convex cases. Second, our approach utilizes labels provided by the users. Those labels can provide important information for us to identify outliers and capture clusters in different densities and complex forms.

\section{Background}\label{pre}
Since SSDBCODI is inspired by the semi-supervised density-based algorithm SSDBSCAN \cite{lelis2009semi}, we will briefly explain some key concepts of density-based clustering in \ref{Pre-Key}, and some of SSDBSCAN's specifics in Section \ref{Pre-algo}.

\subsection{Key Concepts}\label{Pre-Key}
In this section, we introduce some of the most important concepts not just to SSDBSCAN, but to density-based clustering algorithms more widely. The notations used here are illustrated in Table \ref{table:1}. We begin with a dataset, $D \subset \mathbb{R}^{n \times d}$, with $n$ instances and $d$ features.

\begin{table}[h!]
\caption{Notation Table}
\label{table:1}
\centering
\begin{tabular}{||c c c||} 
 \hline
 Notation & Domain & Description \\ [1ex] 
 \hline
 $MinPts$ & $\mathbb{N}$ & Minimum number of points \\ 
 $\epsilon$ & $\mathbb{R}$ & Minimum density threshold \\
 
 $N_{\epsilon}(p)$ & $\mathbb{N}$ & Number of points within $\epsilon$ of $p$ \\
 $D$ & $\mathbb{R}^{n \times d}$ & The whole dataset \\
 $D_{L}$ & $\mathbb{R}^{n_{l} \times d_{l}}$ &The subset of all labeled objects\\
 $D_{O}$ & $\mathbb{R}^{n_{o} \times d_{o}}$ & The subset of labeled outliers\\
 $D_{N}$ & $\mathbb{R}^{n_{n} \times d_{n}}$ & The subset of labeled normal objects\\
 $cDist(p)$ & $\mathbb{R}$ & Core distance of point $p$\\
 $dist(p,q)$ & $\mathbb{R}$ & Distance between $p$ and $q$\\
 $rDist(p,q)$ & $\mathbb{R}$ & Reachability between $p$ and $q$ \\ [1ex] 
\hline

\end{tabular}

\end{table}

The basic assumption of density-based clustering is that clusters are formed with data points located in the dense regions and that those points are defined as core objects. Parameters $\epsilon$ and $MinPts$ are used in density-based clustering to represent the density of a point based on its distance to its neighbouring points. Specifically, if a point is neighboured with at least $MinPts$ number of points within a certain distance $\epsilon$, we determine it as a core object. The following definitions are used to formally define the core object and core distance:
\begin{itemize}
\item \textbf{Definition 1 (Core Object)} A data point $p$ is a core object if $N_{\epsilon}(p) \geq MinPts$. $N_{\epsilon}(q)$ is the number of points within distance $\epsilon$ of point $q$.

\item \textbf{Definition 2 (Core Distance)} $cDist(p)$ is the minimum $\epsilon$ required for a point $p$ to be called a core object. 
\end{itemize}

Data points located in the sparse regions between different clusters are generally considered noise, but these include potential outliers and other unlabeled border points. Since the major step of a density-based clustering algorithm is the expansion from core objects to other core objects, it is important to explore the relationship among core objects as expressed in the following definitions:

\begin{itemize}
\addtolength\itemsep{4mm}
    \item \textbf{Definition 3 (Density-Reachable)} Given $p_{0}, p_{k} \in D$, if there exists a path $P$ from $p_0$ to $p_k$, $(p_{0}$,$p_{1}$,$p_{2}$,...,$p_{k})$ and (1) $\forall p_{i} \in P$, $p_{i}$ is a core object given $MinPts$ and $\epsilon$, and (2) $\forall p_i, p_{i+1} \in P$, $ dist(p_{i},p_{i+1}) \leq \epsilon$,  then $p_{n}$ and $p_{0}$ are density-reachable to each other. In addition, if $p_{0}$ and $p_{n}$ are density-reachable and no point exists between them, then we call them \emph{directly density-reachable}.
    
    \item \textbf{Definition 4 (Reachability)} The reachability  between points $p$ and $q$ is denoted as $rDist(p,q)$ and is calculated by
    \\ $rDist(p,q)=max(cDist(p),cDist(q),dist(p,q))$. 
    \item \textbf{Definition 5 (Density-Connected)} Given points $p, q \in D$, if $\exists v \in D$, such that both $p$ and $q$ are density-reachable from $v$, then $p$ is density-connected to $q$.

\end{itemize}

\begin{figure}[htbp]
\centerline{\includegraphics[width=0.5\textwidth]{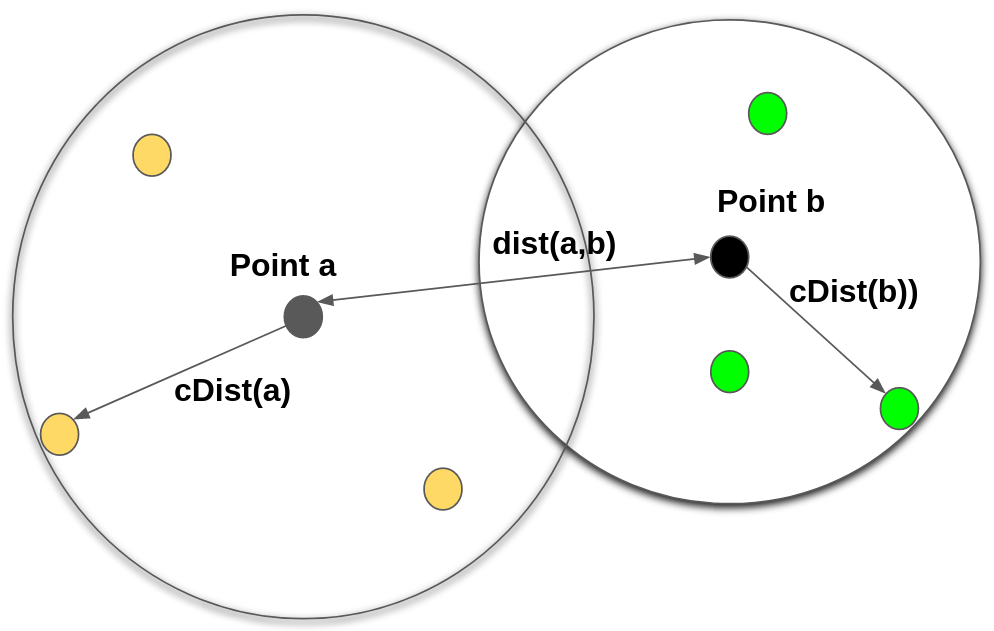}}
\caption{Illustration of cDist and rDist when \textit{MinPts=3}}
\label{fig:reach-dis}
\end{figure}

The relationship between \emph{Reachability} ($rDist$), core distance ($cDist$) and Euclidean distance is illustrated in Figure \ref{fig:reach-dis}, presented as points $a$ and $b$ and given that  $MinPts=3$. Reachability is also an important measure used consistently in the original SSDBSCAN and our proposed approach.

\subsection{Semi-Supervised Density-Based Clustering}\label{Pre-algo}

The idea of SSDBSCAN is to consider all labeled objects as core objects and find optimal $\epsilon$ for each cluster separately instead of using a global $\epsilon$ like the original DBSCAN. An optimal $\epsilon$ should satisfy two conditions: first, all points in the same cluster are density-connected to each other. Second, points of different user labels should not be density-reachable to each other. This ensures that we maximize the cluster size while respecting the not-link constraints given by users.

 \begin{algorithm}
 \caption{SSDBSCAN}
 \begin{algorithmic}[1]\label{algo:1}
 \renewcommand{\algorithmicrequire}{\textbf{Input:}}
 \renewcommand{\algorithmicensure}{\textbf{Output:}}
 \REQUIRE The whole dataset $D$;\\
          Subset of labled objects $D_{L}$
 \ENSURE  Data grouped in clusters $C$
  \STATE $C = \emptyset$
  \FOR {$p \in D_{L}$} 
    \STATE $list = \emptyset$
    \FOR  {$q \in D$}
     \STATE $q.key = \infty$
    \ENDFOR
    \STATE $p.key = 0$
    \STATE $Q = D$
    \WHILE{$Q \ne \emptyset$}
        \STATE $q = $Extract-Min-Reachable($Q$) 
        \STATE $list.add(q)$
        \IF{$q \in D_{L} $ and $p.label \ne q.label$ }
            \STATE $c_{p} =$ BackTrace-Node($list,p$) 
            \STATE $C.add(c_{p})$
            \STATE \textbf{Break}
        \ENDIF
        \FORALL{$s \in Q.adj(q)$}
            \IF{$s \in Q$ and $rDist(q,s) < s.key$}
             \STATE   $s.key = rDist(q,s)$
            \ENDIF
        \ENDFOR
    \ENDWHILE 
  
 \ENDFOR

 \end{algorithmic} 
 \end{algorithm}

The details are illustrated in Algorithm \ref{algo:1}. The algorithm starts with the labeled objects as root points, then from this object $q$, it will iteratively add the next closest object with the shortest $rDist$ to the existing added set. The significance of $rDist$ is given by Lemma 1. 
\begin{itemize}
    \item \textbf{Lemma 1.} The $rDist(p,q)$ measures the minimum $\epsilon$ needed for $p$ and $q$ to be directly density-reachable.
    
  \item \textit{Proof} The proof follows from Definition 4. Let $\epsilon$ = $rDist(p,q)$. Since $rDist(p,q) = max(cDist(p),cDist(q),dist(p,q))$, we have $rDist(p,q) \geq cDist(p)$ and $rDist(p,q) \geq cDist(q)$. This ensures $p$ and $q$ are both core points. In addition, given $rDist(p,q) \geq dist(p,q)$ and Definition 3, we conclude that $p$ and $q$ are directly density-reachable. If $\epsilon < dist(p,q)$, the path containing only two of them violates the conditions of Definition 3 and thus $q$ is not  density-reachable from $p$ under this condition.
\end{itemize}

$rDist$ measures the density between the two core objects. Recall earlier that $\epsilon$ was used as the minimum density threshold for two objects to be density-reachable. Hence, if $rDist$ is relatively low between two core objects, those two objects are more likely to be density-reachable to each other. Given this, among all the neighboring objects, we always visit the one requiring the minimum effort to be directly density-reachable from the currently added set. Then to maximize the cluster size, we record $rDist$ from the next added object to the already-added set, referred to as property \textit{key} in the pseudo-code, and we automatically set the current $\epsilon$ of this cluster to the maximum of all previously recorded $rDist$. This expansion process is similar to Prim's algorithm for constructing the minimum spanning tree (MST) \cite{leiserson1994introduction}. This expansion process also resembles that of the original DBSCAN. According to Corman et al. \cite{ester1996density}, this expansion process ensures a maximality of each cluster under a given $\epsilon$. In other words, as long as a point is density-reachable to the root point w.r.t $\epsilon$, it is guaranteed to be included in the cluster under this expansion process.

\begin{figure}[htbp]
\centerline{\includegraphics[width=0.5\textwidth]{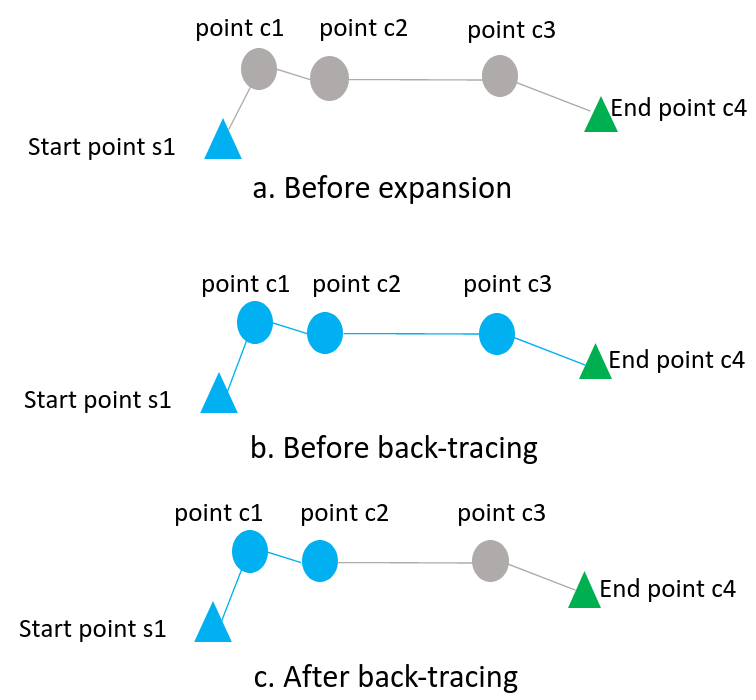}}
\caption{Illustration of the back-tracing process. Triangles represent points labeled by users while the circles represent the originally unlabeled points. The colors illustrate the expansion process.}
\label{fig:back-trace}
\end{figure}

This expansion process terminates when a point $p$ of a different label is met. We back-trace all the added points with their corresponding $rDist$ until we find the maximum edge $E_{max(q,p)}$ along with this expansion path. We also define $p_{max}$ as the end of $E_{max(q,p)}$. After this step, we add all points in the sequence until $p_{max}$ to the same cluster of this root point $q$ before we break the loop and start with a new labeled root point (Lines 12-15). Figure \ref{fig:back-trace} illustrates an example of the back-tracing process, assuming an expansion starts with point $s1$, as depicted in Figure \ref{fig:back-trace}a. In the first step, the expansion starts with $s1$ and terminates when another labeled point c4 is met (Figure \ref{fig:back-trace}b). Then the algorithm will back-trace the longest edge along the path (the edge between $c2$ and $c3$ for this example), and only the points up to $c3$, $c1$, and $c2$ will be included in the cluster of $s1$. $c3$ will remain unclustered after this expansion.

We back-trace the maximum edge  $E_{max(q,p)}$ because the length of this edge represents the minimum density threshold $\epsilon$ required for those points to be density-reachable from the root point $q$ along the expansion path. The proof of this is straightforward as pointed out in Lelis et al. \cite{lelis2009semi}. Earlier, we have defined $E_{max(q,p)}$ as the value of the longest edge length along the expansion path from $q$ to $p$. Suppose we set the actual $\epsilon < E_{max(q,p)}$, then the expansion from the root point $q$ will terminate at this longest edge before it actually reaches $q$. Because of this, $p$ will never be included in the same cluster with $q$. Conversely, if we set $\epsilon \geq E_{max(q,p)}$, $p$ is guaranteed to be density-reachable from $q$. 

Therefore to avoid a point of a different label being wrongly included in the cluster, we set $\epsilon$ to be a little lower than $E_{max(q,p)}$, which is equivalent to adding all points up to $p_{max}$. Lines 17-19 visit the adjacent nodes and update their edge values, similar to the standard Prim's algorithm.
\par

Overall, this algorithm has two main advantages. First, since we do not set a global $\epsilon$, each cluster is gradually learning its optimal $\epsilon$ through the expansion process, helping to differentiate clusters of different densities. Second, the back-tracing mechanism ensures that $rDist$ is self-adjusted to conform to the not-link constraints suggested by the user labels.

\section{SSDBCODI}\label{method}
The SSDBSCAN algorithm is superior in dealing with clusters of different densities, while using back-tracing ensures the clusters do not violate the provided labels (see Section \ref{pre}). However, the back-tracing process often leaves some points unclustered, as point $c3$ illustrated in Figure \ref{fig:back-trace}.

In the original SSDBSCAN, after back-tracing, the algorithm assigns the unclustered points to their nearest core points, assuming that every point belongs to at least one of the commonly identified clusters. However, as discussed in Sections \ref{Intro} and \ref{related}, such an assumption does not consider the existence of outliers, which could belong to none of the common clusters, negatively affecting the overall clustering process. 

Spurred by this challenge, we propose a semi-supervised density-based solution that provides an effective mechanism to detect and actively remove the outliers during the clustering process. This section introduces our proposed outlier detection approach and then illustrates how we incorporate the outlier detection into a semi-supervised classification to cluster the unclustered points left out by the back-tracing process.

\subsection{Outlier Detection} \label{outlier-detect}
Outliers coexist with normal data points. To effectively differentiate them, certain assumptions should be applied to describe outliers. We apply the following assumptions, which utilize the $rDist$ defined in Section \ref{pre} and $dist$ that refers to the Euclidean distance between the  instances reported in Table 1: 

\begin{itemize}

    \item \textbf{Assumption 1. (Reachability Assumption)} The outliers are less density-reachable from all normal clusters, whereas normal points are relatively more density-reachable to at least one normal cluster.
    
    \item \textbf{Assumption 2. (Local Density Assumption)} The outliers are usually located in relatively sparse regions compared to normal points.
    
    \item \textbf{Assumption 3. (Similarity Assumption)} Certain outliers are usually closer to each other than to normal points \cite{zhang2018anomaly}.
\end{itemize}

In Section \ref{pre}, we explained the idea of back-tracing, and its contribution to finding the minimum $\epsilon$ for points $p$ and $q$ to be density-reachable and thus belong in the same cluster. This value is noted as $E_{max(p,q)}$. In other words, for a labeled point $p$, if we set $\epsilon < E_{max(p,q)}$, $q$ will never be reached during the expansion. Therefore $q$ will never be included in the same cluster. This idea was originally used to separate points of different clusters and proved effective by Lelis et al.\cite{lelis2009semi}. Since the outliers are different from normal clusters, a potential outlier should be relatively less density-reachable to all labeled normal instances compared to the normal point. This conclusion can be used for outlier detection, and Assumption 1 is the natural extension of this concept. To formally quantify this value, we provide the following formulas, assuming $q$ is an arbitrary point and $p_{i} \in D_{N}$ is the $i^{th}$ labeled normal object:
 \begin{equation}\label{eq:1}
      E_{max(q, p_{min})}=min_{s \in D_{N}}\{E_{max(q, s)}\}
\end{equation}

\begin{equation}\label{eq:2}
      rScore(q)=e^{-E_{max(q,p_{min})}}
  \end{equation}

\noindent In Equation \ref{eq:1}, we aim to find the labeled normal instance $p_{min}$, which refers to the most density-reachable labeled normal point to $q$. For the outliers we expect their value to be significantly larger than the normal instances. Equation \ref{eq:2} is simply normalizing  $E_{max(q,p_{min})}$, so its range is$[0,1]$. This value is defined as the \textit{Reachability-Score} noted as $rScore(q)$, which is used as a property for $q$, where a lower $rScore(q)$ leads to greater possibility to be an outlier for $q$. \par
Apart from the $rScore$, we also introduce two other indicators based on Assumptions 2 and 3: the \textit{Local -Density-Score} and the \textit{Similarity-Score}. The former, denoted as $lScore$, measures the sparsity of a given point around its neighboring regions, while the latter, denoted as $simScore$, measures how close a point is to the nearest labeled outlier. Assuming $nearest_i(q)$ refers to the $i^{th}$ nearest neighbour to $q$ according to $rDist$, the indicators are computed as follows:

\begin{equation}\label{eq:local density}
LD(q)=\frac{1}{MinPts}\mathlarger{\mathlarger{\sum}}_{i=1}^{MinPts}rDist(nearest_i(q),q)
\end{equation}
\begin{equation}\label{eq:sim-dist}
    dist(q,o_{min})=min(dist(q,o_{i}) \mid \forall o_{i} \in D_{O}\})
\end{equation}
\begin{equation}\label{eq:local score}
    lScore(q)=e^{-LD(q)}
\end{equation}
\begin{equation}\label{eq:sim-score}
    simScore(q)=e^{-dist(q,o_{min})}
\end{equation}
Equation \ref{eq:local density} reflects Assumption 2 of the local density. For a given point $q$, we compute its local density by averaging its neighbouring $rDist$ to $q$. The number of neighbours chosen to compute this average value is optional based on people's expertise and experience. Here we choose $MinPts$ to be consistent with the $rDist$ computation. Equation \ref{eq:sim-dist} aims to reflect Assumption 3 by computing the distance from $q$ to its nearest labeled outlier, denoted as $o_{min}$ in the equation. Similarly to Equation \ref{eq:2}, Equations \ref{eq:local score} and \ref{eq:sim-score} are normalization transformations to convert $LD(q)$ and $dist(q,o_{min})$ with the $[0,1]$ range. Since the three scores (rScore, lScore, simScore) detect outliers based on different aspects, we enhance robustness by calculating their weighted average:
\begin{equation}\label{eq:total}
    \begin{split}
          tScore(q)=\alpha (1-rScore(q))  \\+ \beta (1-lScore(q)) \\ +(1-\alpha-\beta)simScore(q)\\
          where \ \alpha,\beta \in [0,1] \ and \ \alpha +\beta \leq 1
    \end{split}
\end{equation}
Equation \ref{eq:total} refers to the weighted average of the three computed outlier detection scores. In this equation, $\alpha$ and $\beta$ are two hyper-parameters representing the importance assigned to each score before training. The terms $(1-rScore(q))$ and $1-lScore(q)$ are simply reversing $rScore(q)$ and $lScore(q)$ so higher scores indicate a higher probability for point $q$ to be an outlier.

\subsection{Semi-Supervised Classification}\label{classification}
Initially, we run SSDBSCAN based on the initial labeled points and $rScore$, $lScore$, and $simScore$ are computed and then averaged to calculate the $tScore$. In the meantime, some points are clustered, which for convenience, are noted as reliable normal points $R_{N}$. As already explained, $R_{N}$ are clustered using SSDBSCAN and refer to the core objects that are mostly density-reachable to the labeled points. Then the remaining points are ordered based on $tScore$, and top $k$ outliers are selected as reliable outliers $R_{O}$, where $k$ is a user-defined hyper-parameter. However, utilizing back-tracing when running the SSDBSCAN imposes certain points to remain unclustered (Figure \ref{fig:back-trace}). To further cluster these points, instead of simply assigning them to their nearest core points, as suggested by the original SSDBSCAN paper, we apply a non-linear classifier trained using $R_{N}$ and $R_{O}$ as training labels. This strategy affords a non-linear classifier to create clusters of complex and non-convex shapes. During the training process,  $rScore$ and $tScore$ are used as the instance weights for $R_{N}$ and $R_{O}$, respectively. 

 \begin{algorithm}
 \caption{Proposed Algorithm}
 \begin{algorithmic}[1]\label{algo:2}
 \renewcommand{\algorithmicrequire}{\textbf{Input:}}
 \renewcommand{\algorithmicensure}{\textbf{Output:}}
 \REQUIRE        
          The whole dataset $D$;\\
          Subset of labled normal objects $D_{N}$;\\
          Subset of labeled outliers $D_{O}$;\\
          Number of reliable outliers $k$
 \ENSURE  Predicted normal clusters $C_{n}$;\\
          Predicted Outliers $C_{o}$
 \STATE Run Algorithm \ref{algo:1} using $D$, $D_{N}$ and $D_{O}$ 
 \STATE Compute $tScore$ and $rScore$; Collect $R_{N}$ and $R_{O}$
 \STATE Train a classifer $H$ using $tScore$, $rScore$, $R_{N}$ and $R_{O}$
 \STATE Predict $C_{n}$ and $C_{o}$ using $CL$ on $D$

 \end{algorithmic} 
 \end{algorithm}
 
\begin{figure}[htbp]
\centerline{\includegraphics[width=0.5\textwidth]{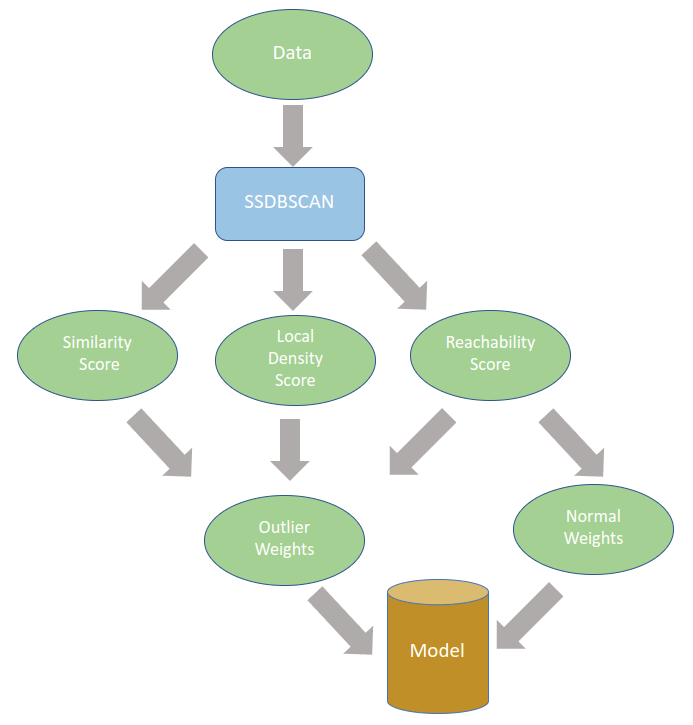}}
\caption{Summary of the entire process}
\label{fig:entire-process}
\end{figure}

 Algorithm \ref{eq:2} introduces the steps of our proposed algorithm.
 Considering implementation, Lines 1 and 2 can be executed simultaneously, and according to Lelis and Sander  \cite{lelis2009semi}, the time complexity of this process is $O(nN^2lgN)$, where $N$ is the number of cases in the entire dataset, and $n$ is the number of normal labeled objects. Also, several minor modifications should be made to Algorithm \ref{algo:1} to fit our purposes:
 \begin{itemize}
     \item In Line 15 of Algorithm \ref{algo:1}, we break the loop when a point with a label differs from the root point that is met during the expansion. However, the proposed algorithm does not terminate at this point because $rDist$ must be computed for every object to calculate the $rScore$ at a later stage. However, we still run the back-trace function in Line 13 to record the $RN.$
     \item When running the Algorithm \ref{algo:1}, we only expand from the labeled normal points $D_{N}$ since $rScore$ is only computed for $D_{N}.$
     \item We treat the labeled outliers $D_O$ as different classes from all normal labels. Therefore the Trace-Max function will also run when the expansion meets a labeled outlier point during Algorithm \ref{algo:1}.
 \end{itemize}
 Figure \ref{fig:entire-process} summarizes the process of the proposed algorithm.

\section{Evaluations}\label{eval}
To analyze the performance of our proposed algorithm,
we have run experiments against real-world data. Although our algorithm performs both clustering analysis and outlier detection at the same time, in order to have a more detailed analysis of our algorithm, evaluations on clustering and outlier detection are performed independently.

\subsection{Evaluation Metrics} \label{eval-metric}
\begin{figure}[htbp]
\centerline{\includegraphics[width=0.53\textwidth]{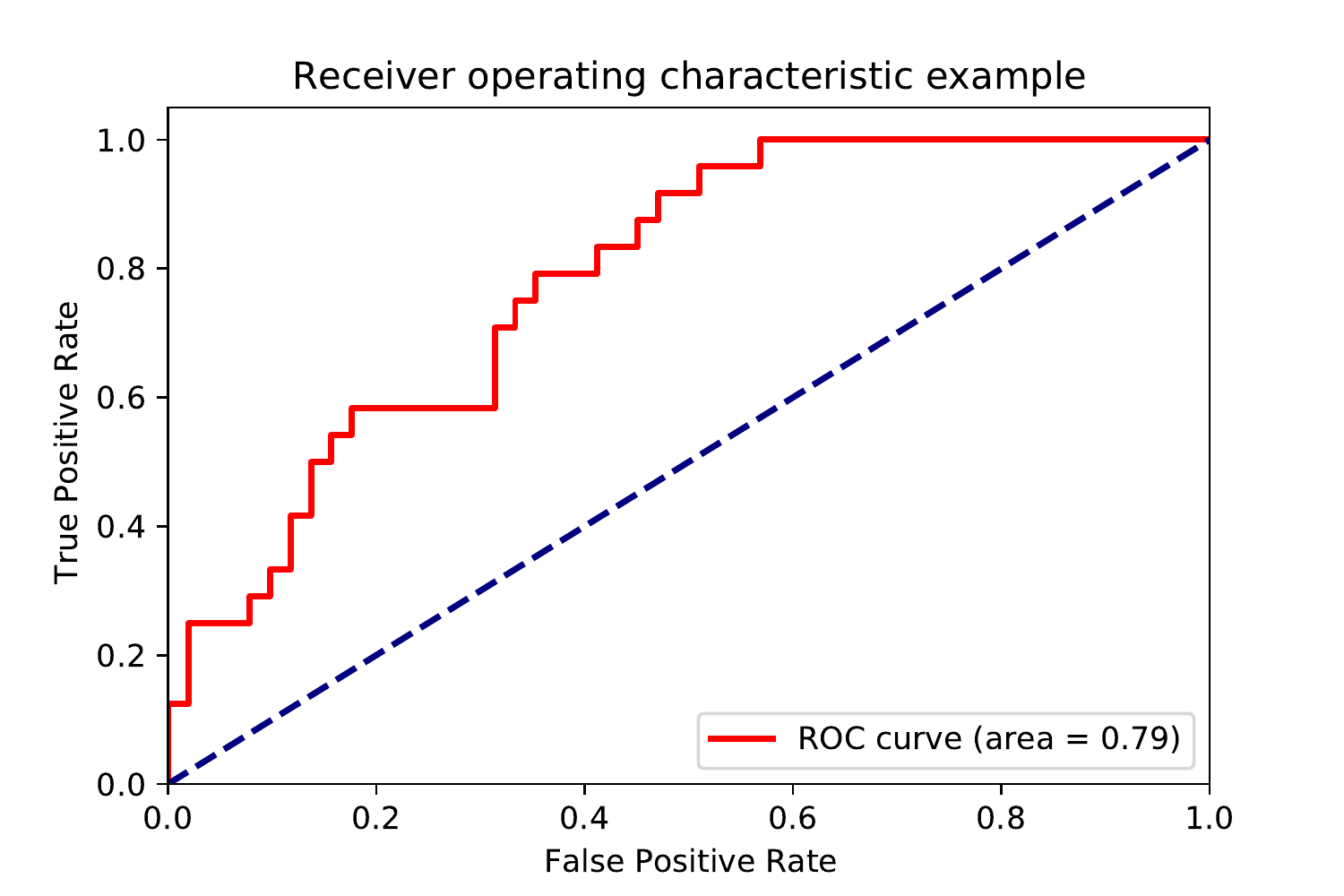}}
\caption{Example of a typical ROC Curve 
}
\label{fig:roc-example}
\end{figure}

For outlier detection, we use \emph{area under the curve} noted as AUC score. This metric is widely used in outlier detection across various research fields \cite{hodge2004survey,chandola2007outlier,wang2019progress,boukerche2020outlier}. The AUC score is in general computed from the \emph{receiver operating characteristic} curve (ROC curve) as illustrated with a typical example in Figure  \ref{fig:roc-example}.
The x-axis represents the false positive rate (FPR, the number of incorrectly predictive positive cases divided by the total number of negative cases) and the y-axis represents the true positive rate (TPR, number of correctly predicted positive cases divided by the total number of positive cases). The ROC curve shows the tradeoff between these quality measures as the thresholds or parameters of an algorithm are adjusted. The AUC is simply the area of the region under the ROC curve, i.e.~under the red line shown in Figure \ref{fig:roc-example}. Though an ROC curve can be used to manually tune a model's parameters, in the context of evaluation with AUC, the curve is automatically generated with a sequence of thresholds.  A higher AUC in general indicates that the classification model could achieve a relatively high TPR without tolerating a significant FPR. With $AUC = 0.5$ (the blue line in the figure) the classification is as good as a random guess.

For clustering, Rand Index and Normalized Mutual Information (NMI) are used for the evaluation. Those measures are two of the most used measures for clustering analysis according to Wu et al. \cite{wu2009adapting}.

Rand Index noted as \textit{R} proposed by William M. Rand \cite{rand1971objective} measures the agreement between two clusters given the same data with size \textit{n} as illustrated in Equation \ref{eq:rand-index}.
\begin{equation}\label{eq:rand-index}
   R= \frac{a+b}{\binom{n}{2}}
\end{equation}
In this equation, \textit{a} represents the number of pairs of data instances assigned to the same cluster while \textit{b} represents the number of pairs of instances assigned to different clusters in both the predicted sets and the ground-truth labels. This is a widely used measure for evaluation on clustering performance and is consistent with the metric used in the original SSDBSCAN.

Normalized mutual information (NMI) is a normalized version of mutual information, an information theory metric which evaluates how much a given random variable tells about another variable. In our case, those two variables refer to the ground-truth labels and the predicted clustering labels. 
\begin{equation}\label{eq:nmi}
   NMI = \frac{2 \times I(Y;C)}{H(Y)+H(C)}
\end{equation}
In the equation, $Y$ and $C$ refer to the variables formed by the ground-truth labels and predicted clustering labels respectively. $I(Y;C)$ refers to the mutual information between $Y$ and $C$. $H(Y)$ and $H(C)$ refer to the entropy of $Y$ and $C$ respectively.

\subsection{Experiment Settings}
We use six different datasets for evaluation. Table \ref{table:data} illustrates the key characteristics of those datasets. 

\begin{table}[h!]
\caption{Key Characteristics of Datasets}
\label{table:data}
\centering
\begin{tabular}{||c c c c c||} 
 \hline
 Data Name & \#instances & \#attributes & \#outliers & \#clusters \\ [1ex] 
 \hline
 
 $lympho$ & 148 & 18 & 6 & 2\\ 
 $ecoli$ & 336 & 7 & 9 & 5 \\
 $arrhythmia$& 452 & 274 & 66 & 4\\
 $yeast$ & 1484 & 8 & 185 & 4\\
 $satellite$ & 6435 & 36 & 2036 & 3\\
 $pendigits$ & 6870 & 16 & 156 & 9\\ [1ex]
\hline
\end{tabular}
\end{table}

\begin{figure*} 
\begin{multicols}{3}
    \includegraphics[width=\linewidth]{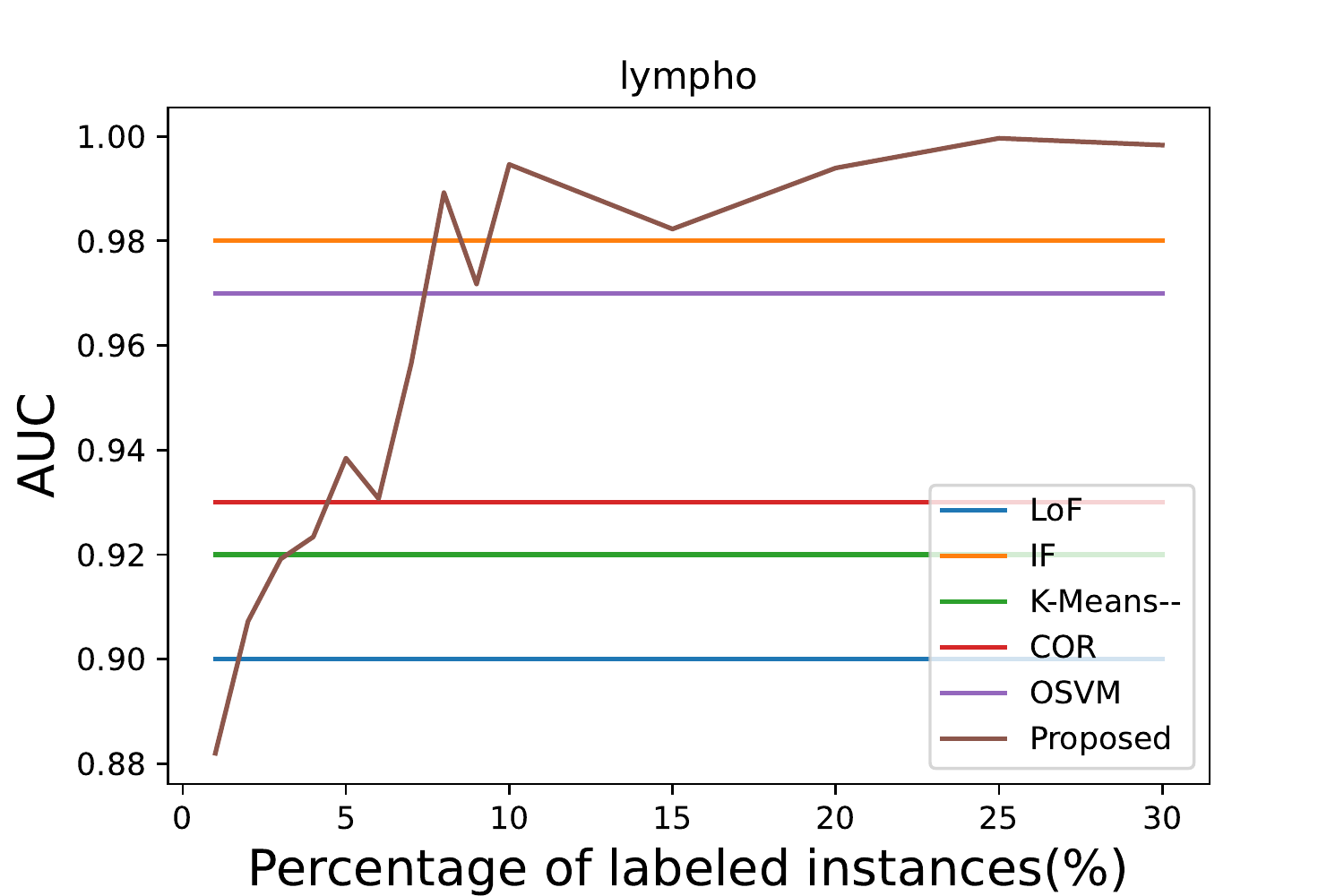}\par 
    \includegraphics[width=\linewidth]{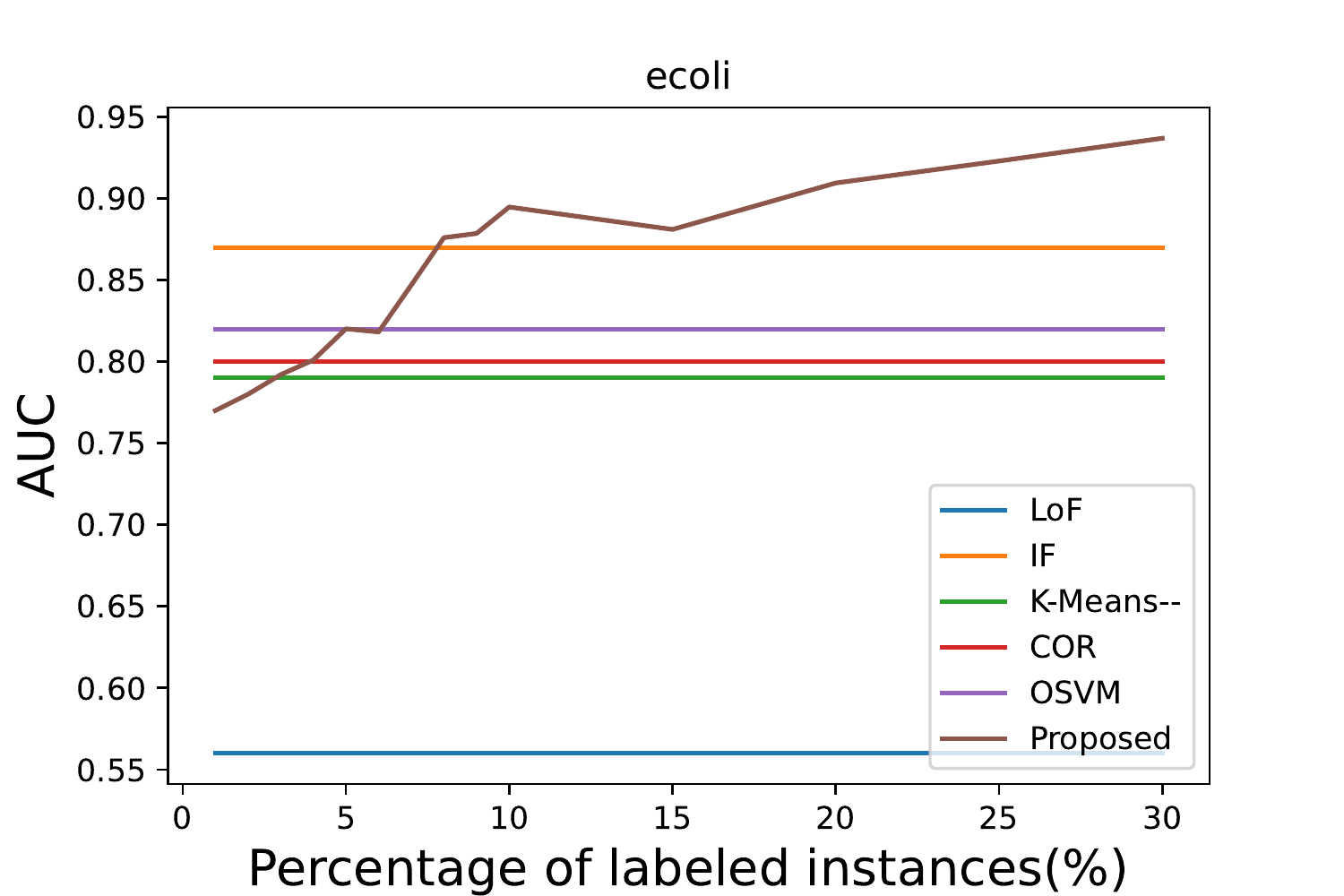}\par 
    \includegraphics[width=\linewidth]{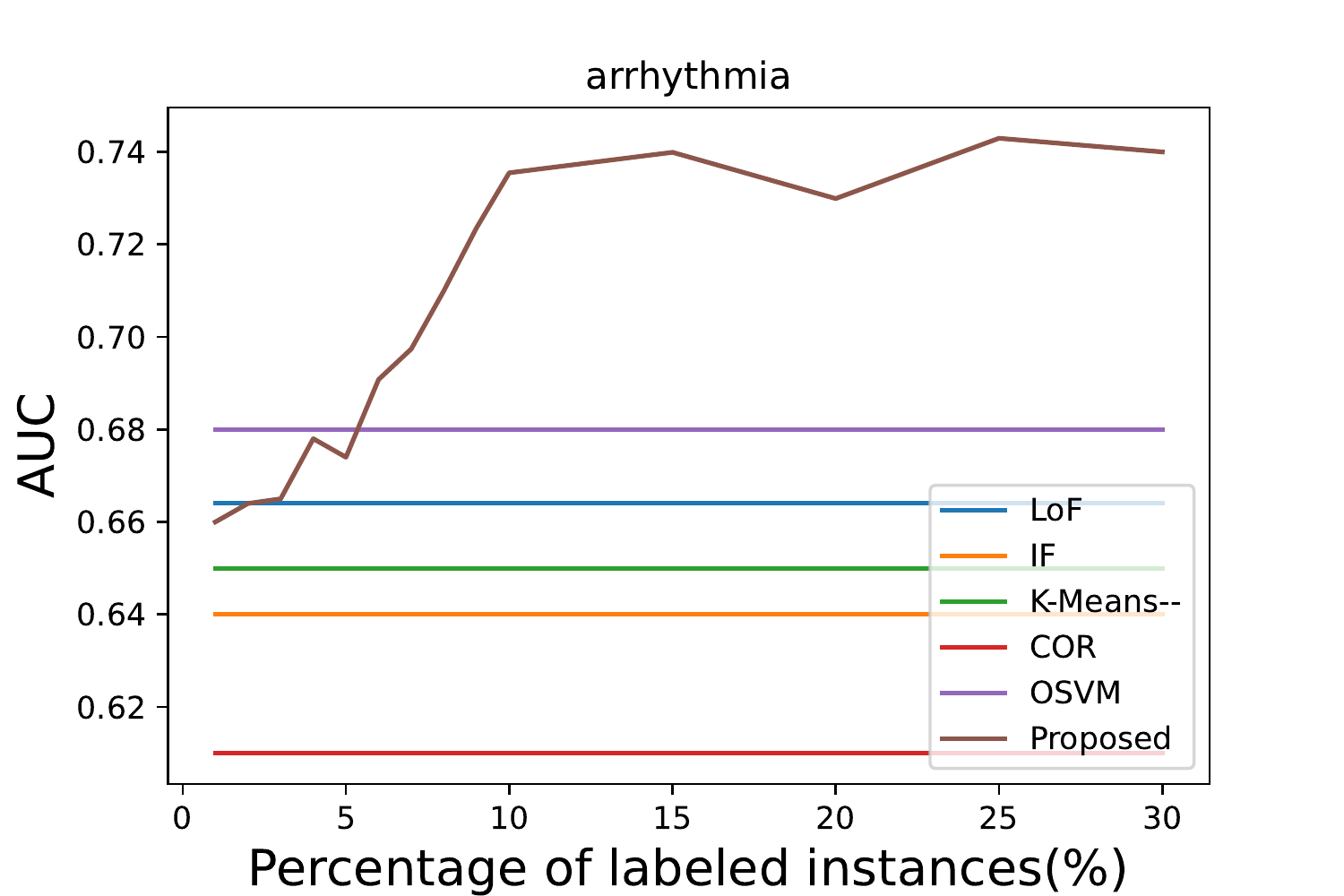}\par
    \end{multicols}
\begin{multicols}{3}
    \includegraphics[width=\linewidth]{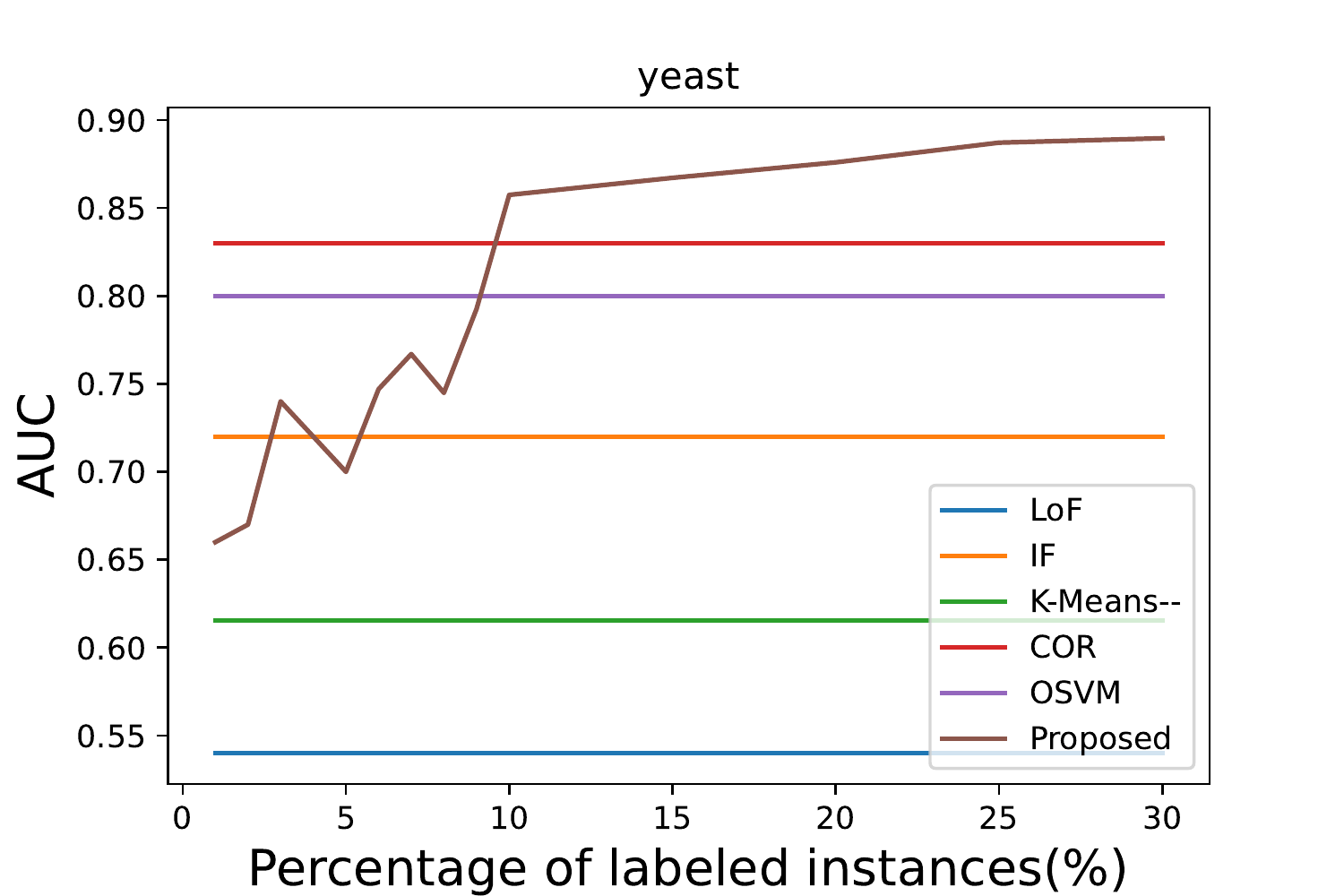}\par
    \includegraphics[width=\linewidth]{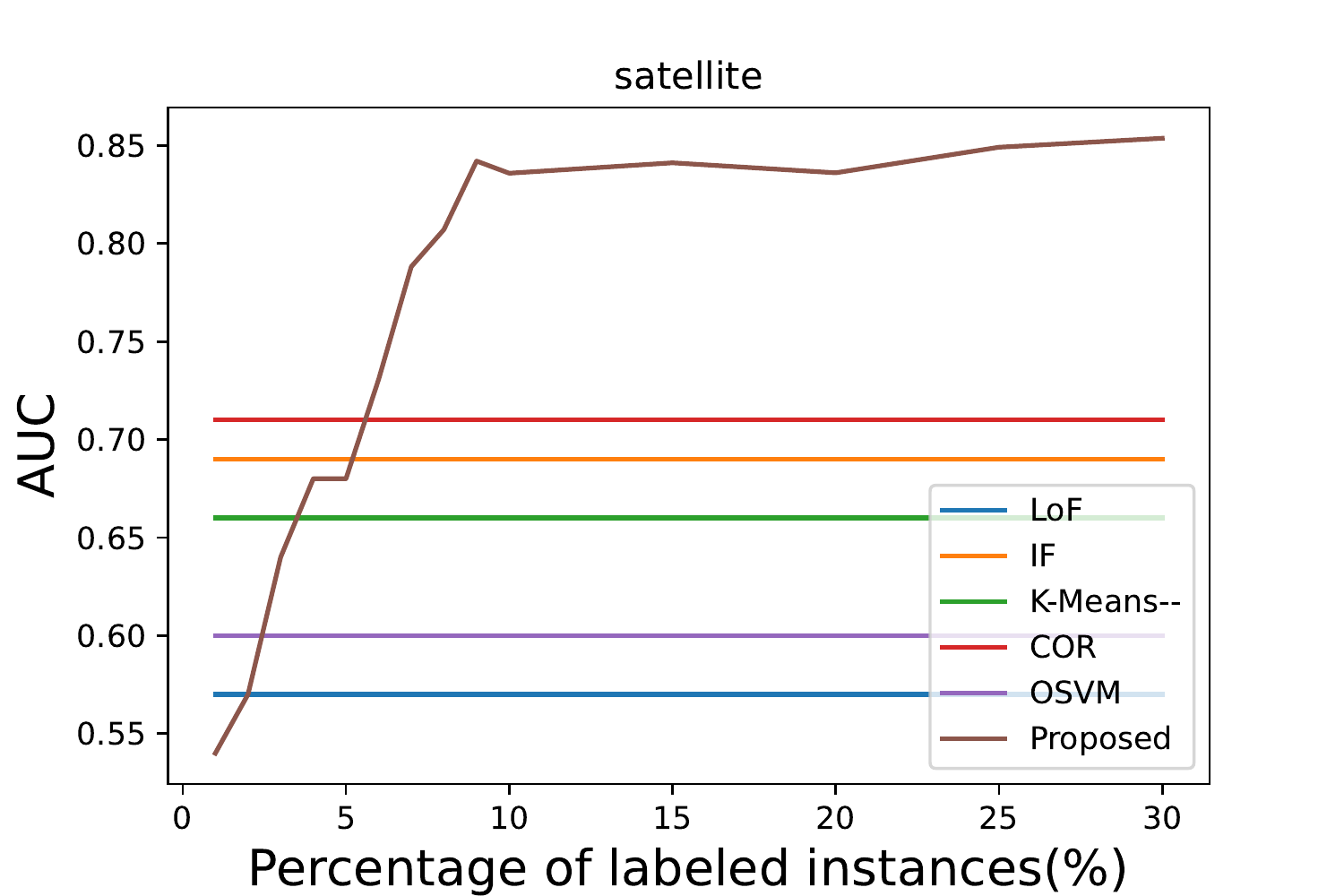}\par
    \includegraphics[width=\linewidth]{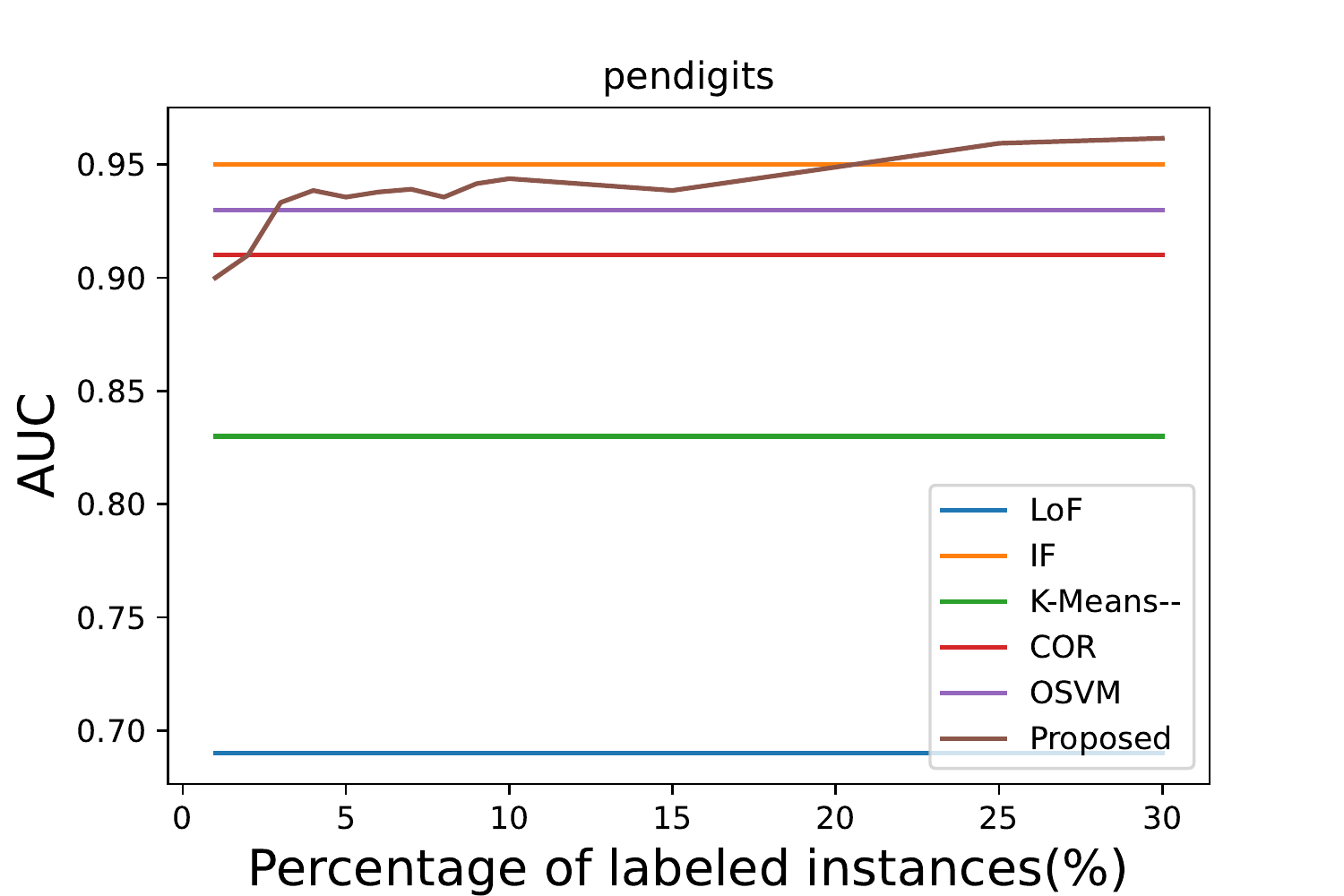}\par
\end{multicols}
\caption{Evaluation Results of the Outlier Detection}
\label{fig:eval-out}
\end{figure*}

\begin{figure*}

\begin{multicols}{3}
    \includegraphics[width=\linewidth]{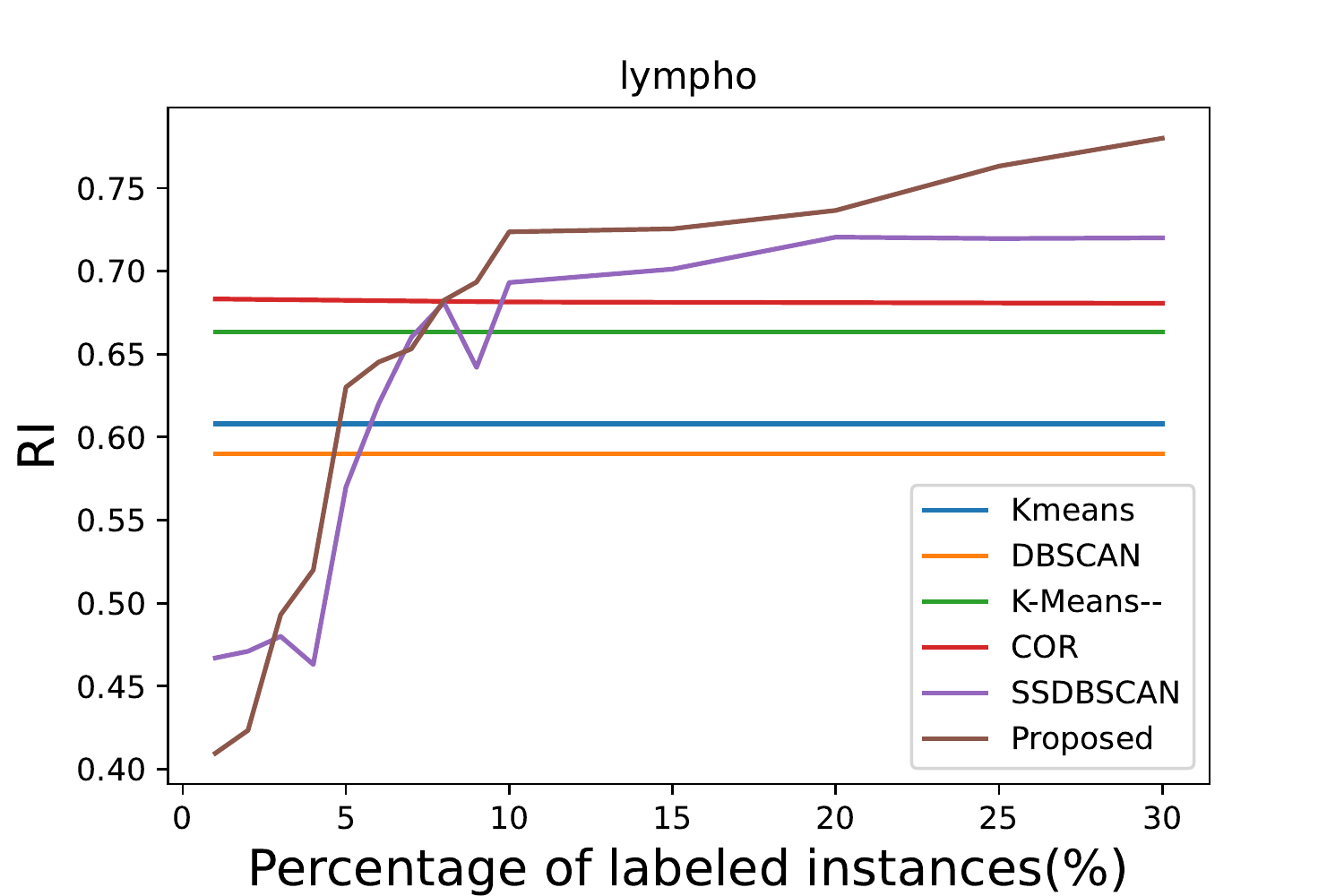}\par 
    \includegraphics[width=\linewidth]{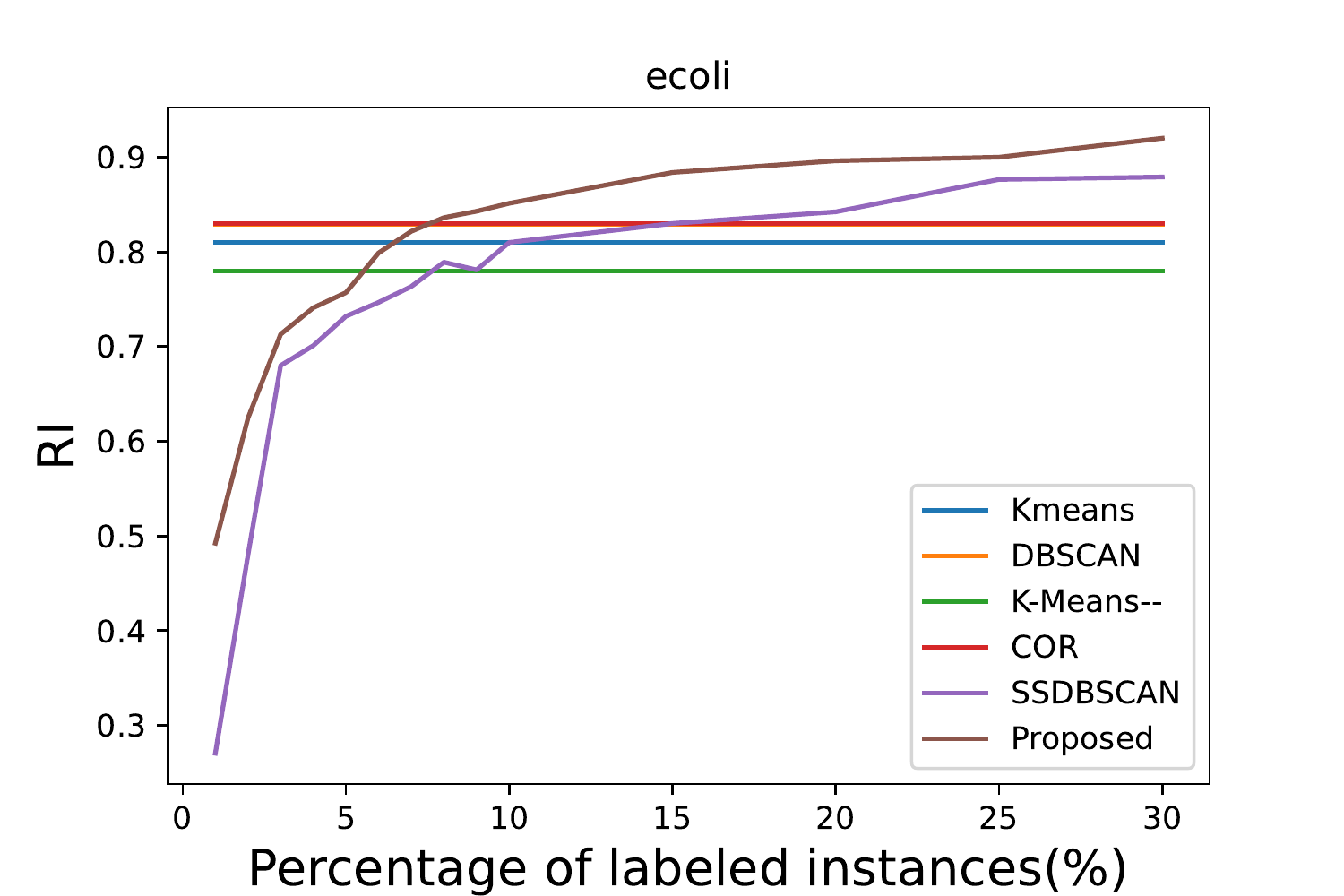}\par 
     \includegraphics[width=\linewidth]{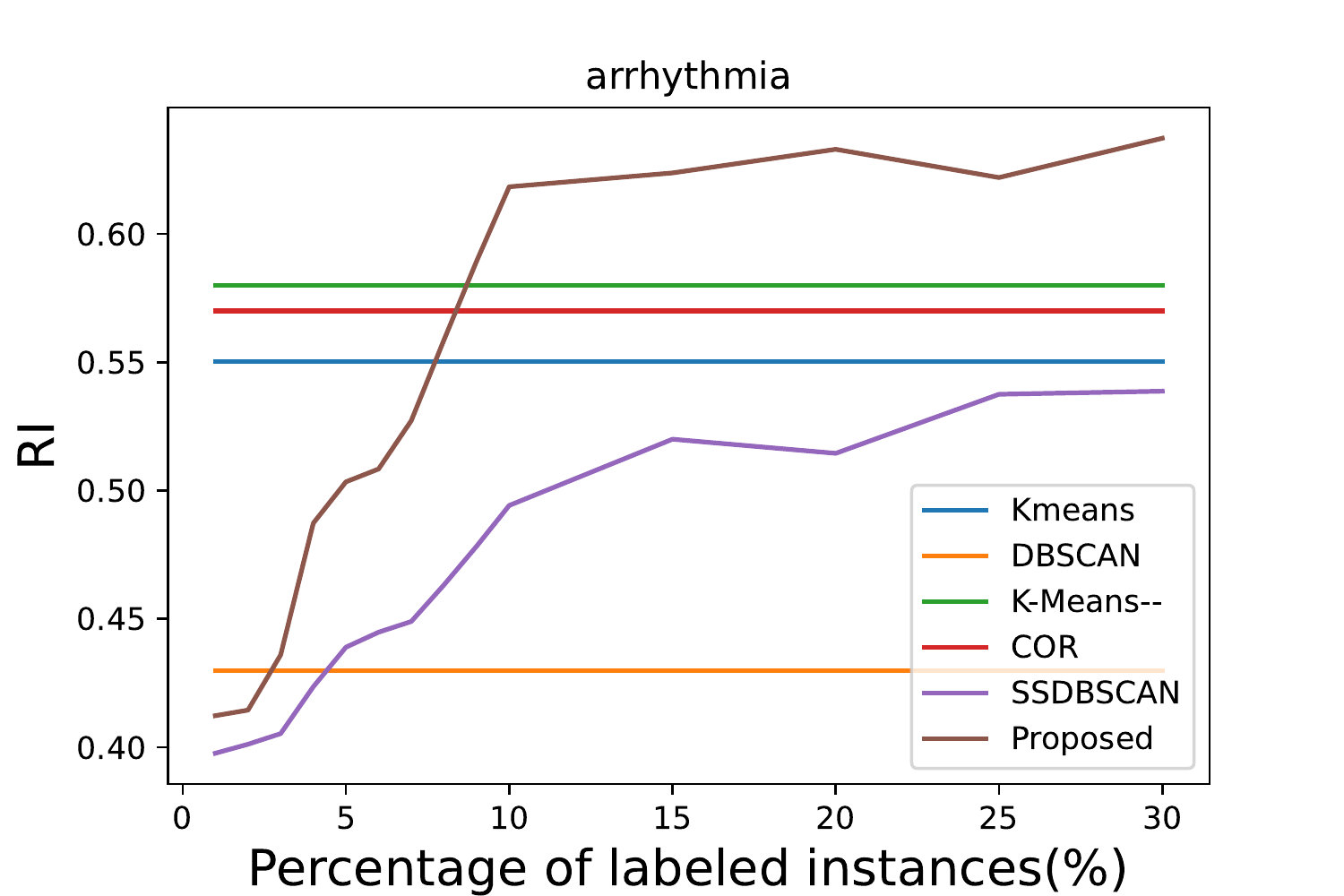}\par
    \end{multicols}

\begin{multicols}{3}
    \includegraphics[width=\linewidth]{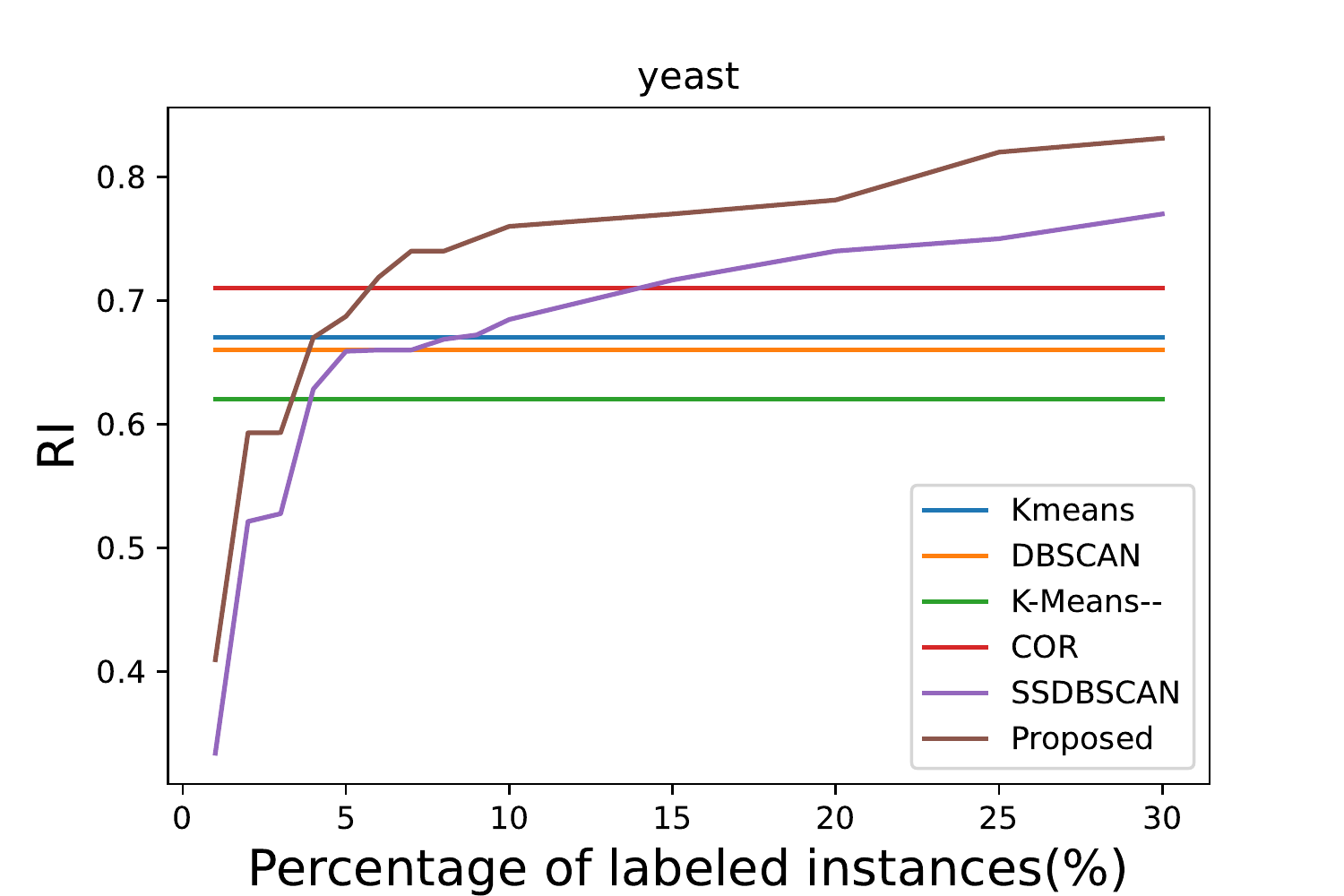}\par
    \includegraphics[width=\linewidth]{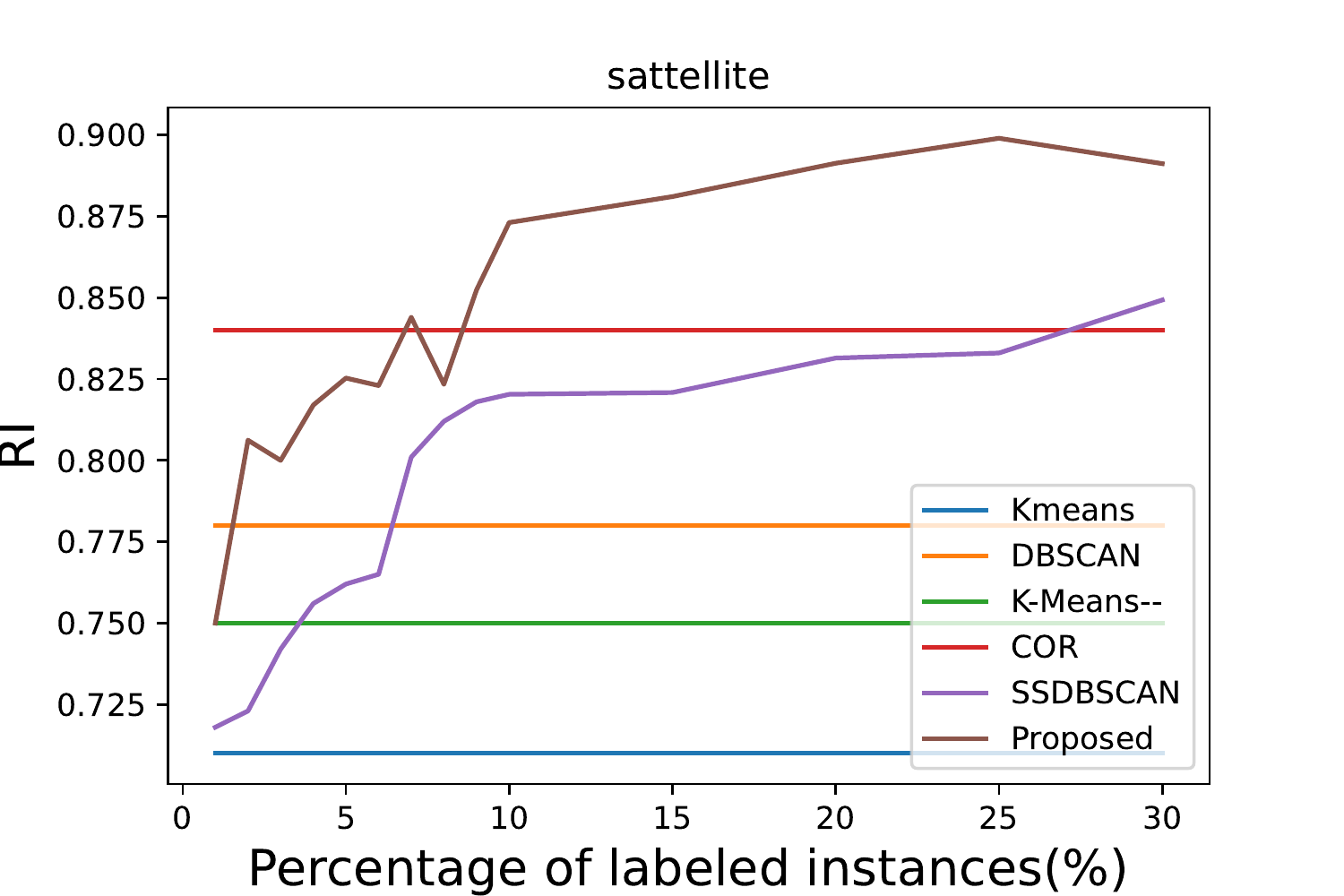}\par
    \includegraphics[width=\linewidth]{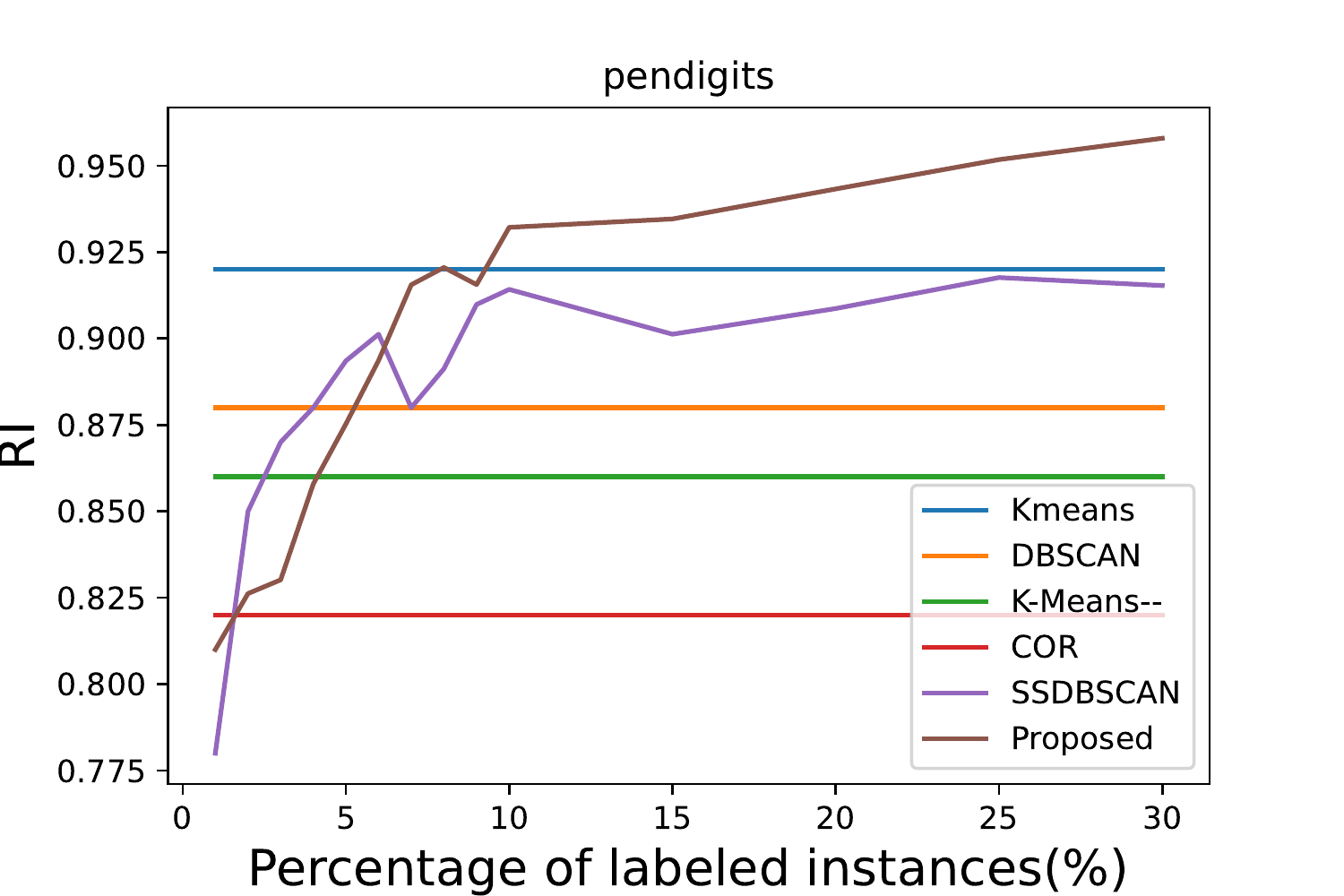}\par

\end{multicols}
\caption{Evaluation Results of the Clustering in Rand Index Score}
\label{fig:eval-cluster}
\end{figure*}

\begin{figure*}

\begin{multicols}{3}
    \includegraphics[width=\linewidth]{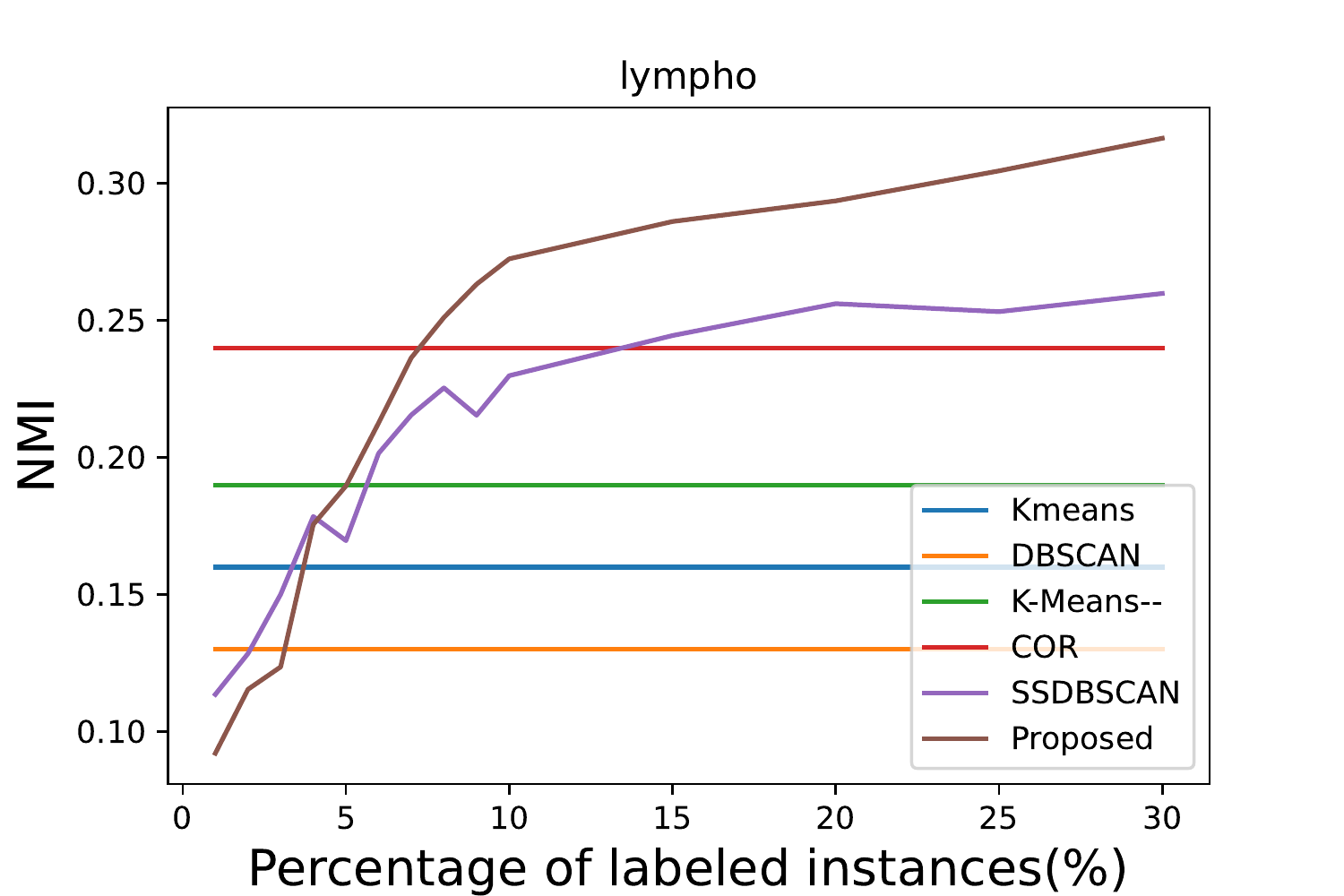}\par 
    \includegraphics[width=\linewidth]{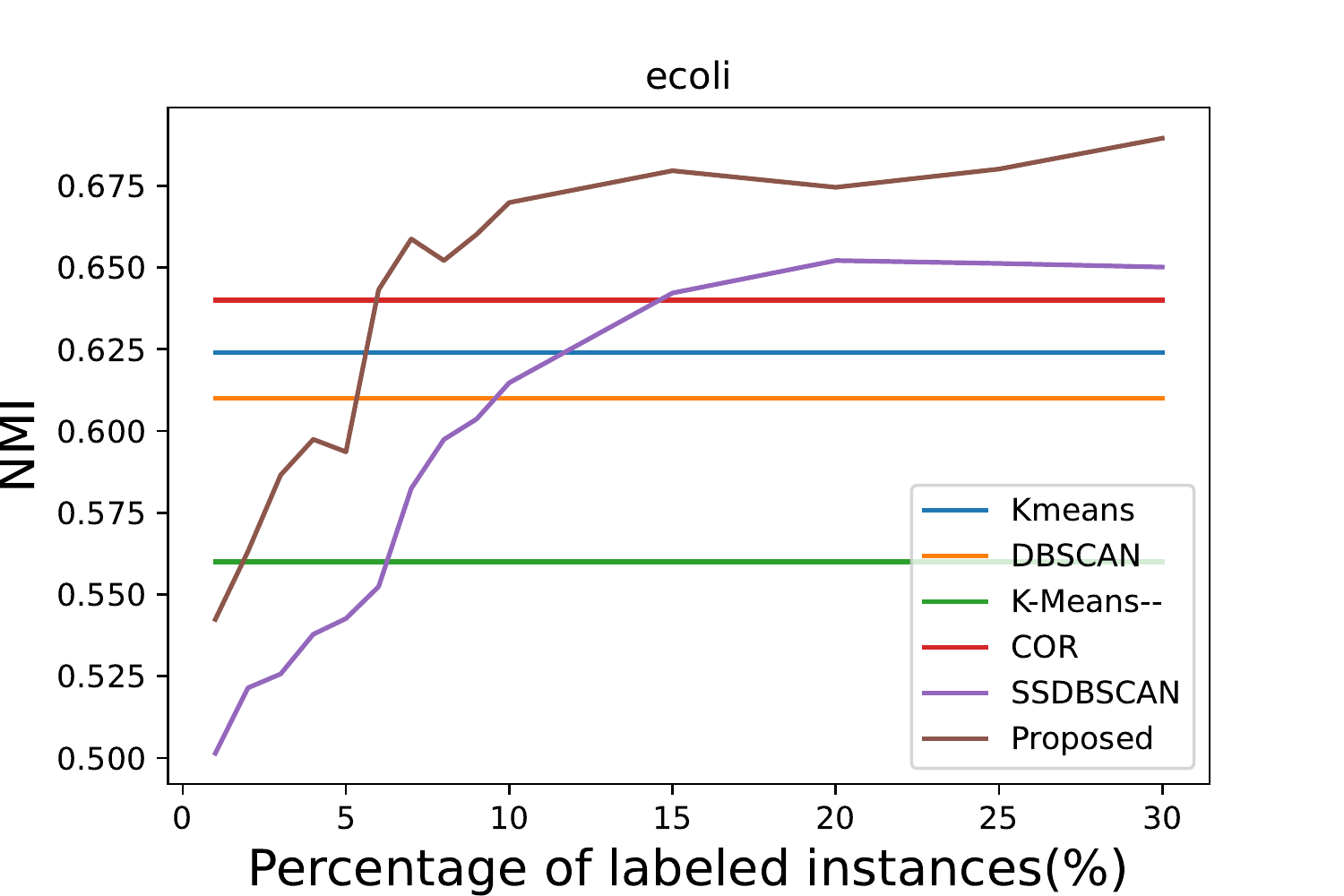}\par 
     \includegraphics[width=\linewidth]{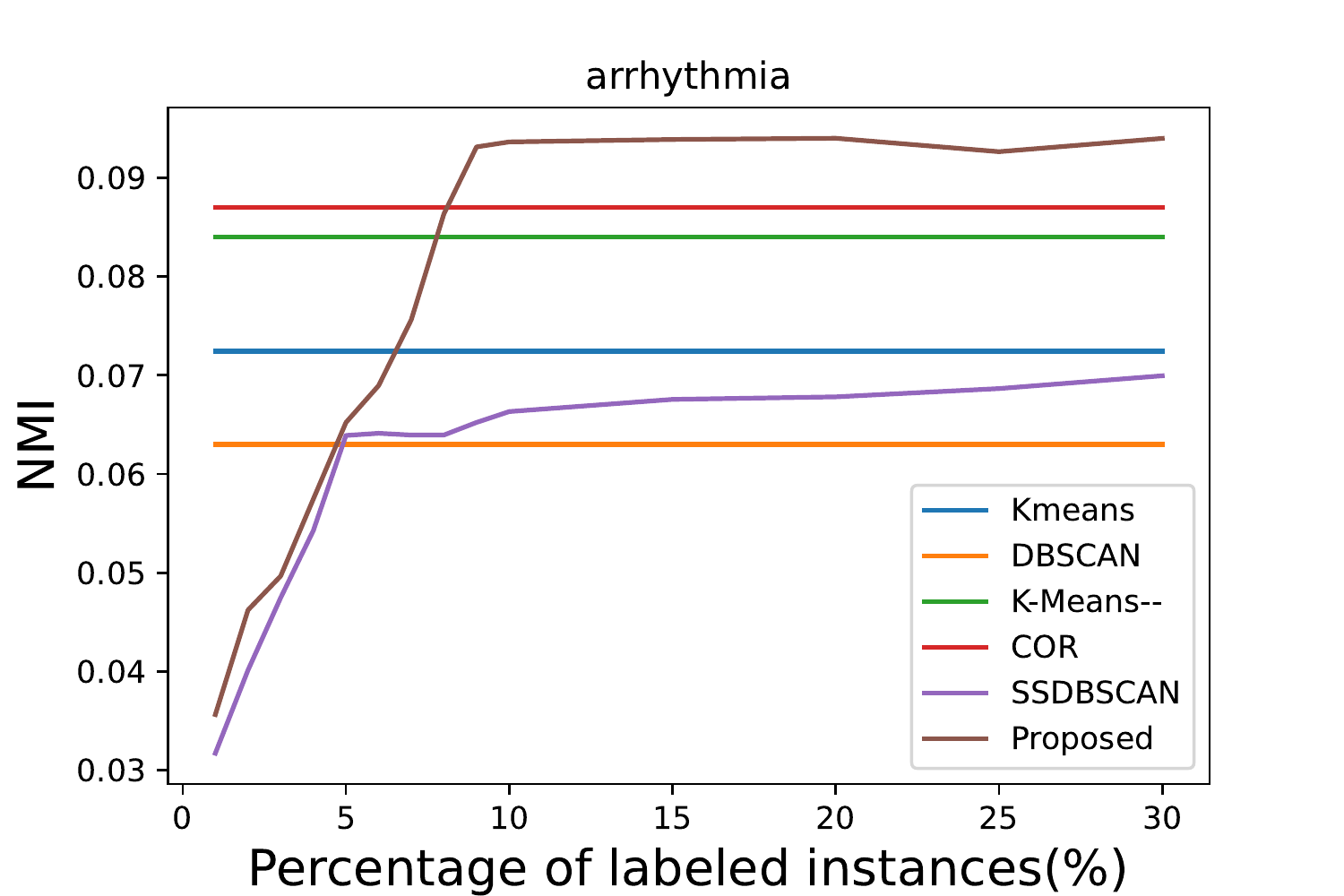}\par
    \end{multicols}

\begin{multicols}{3}
    \includegraphics[width=\linewidth]{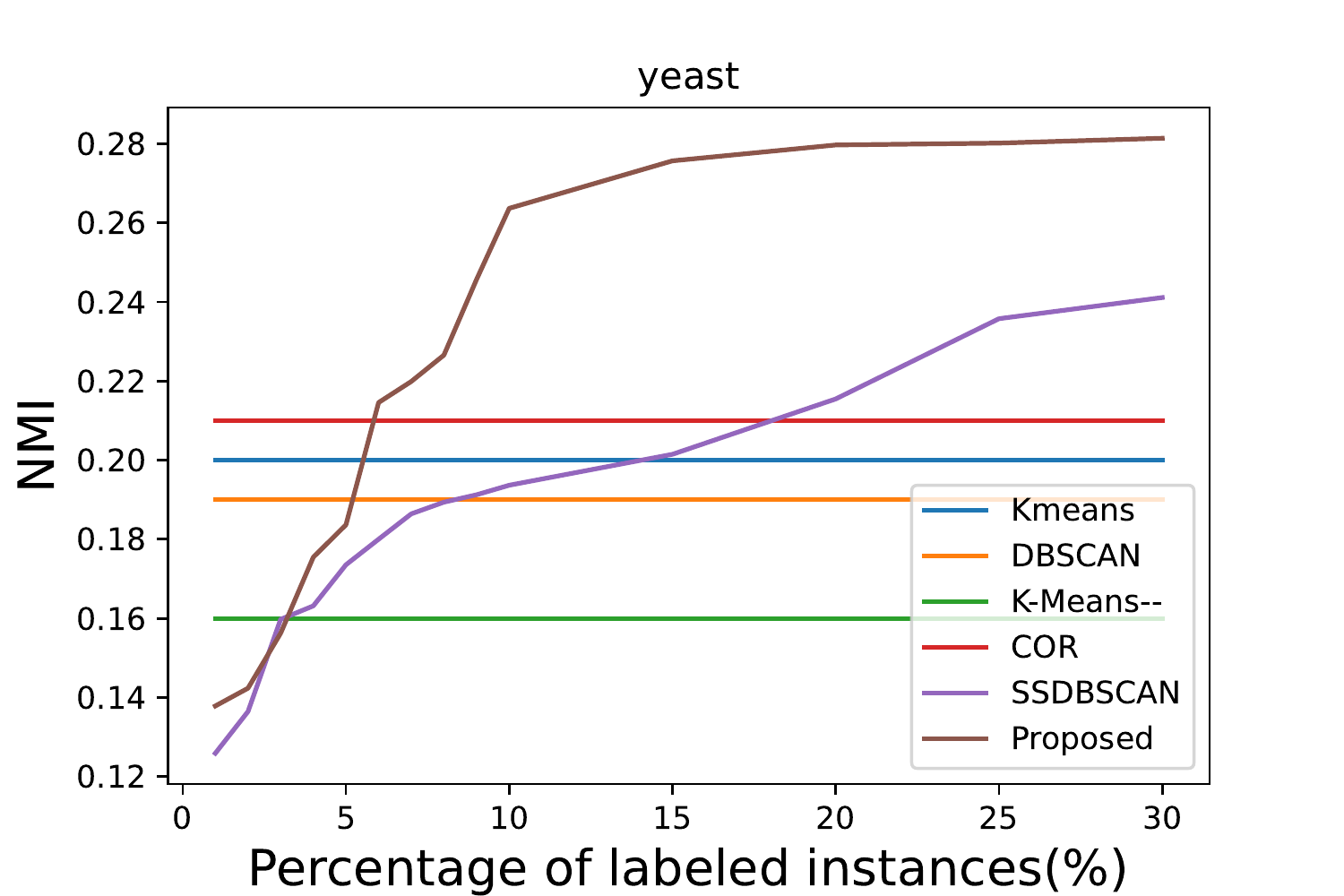}\par
    \includegraphics[width=\linewidth]{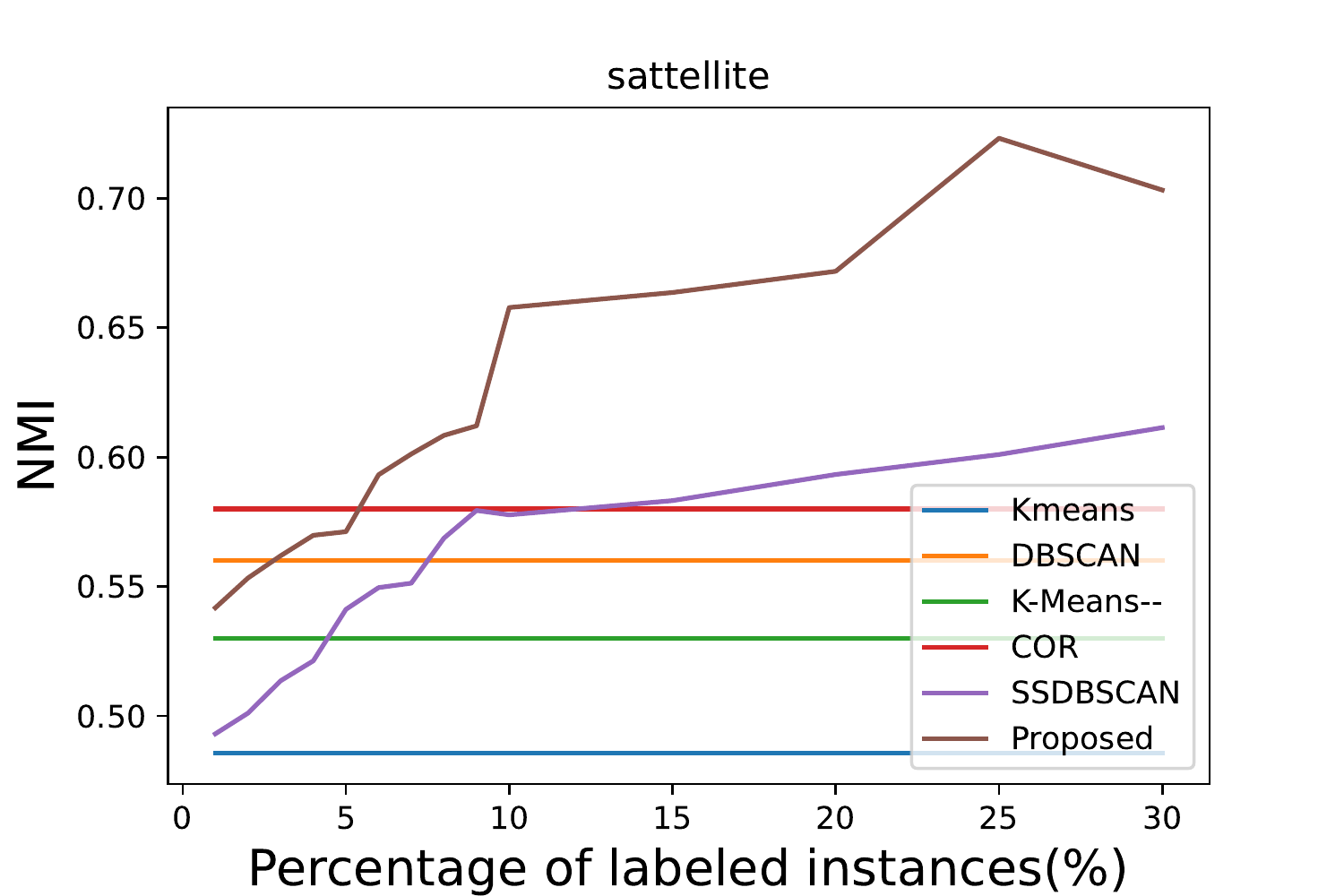}\par
    \includegraphics[width=\linewidth]{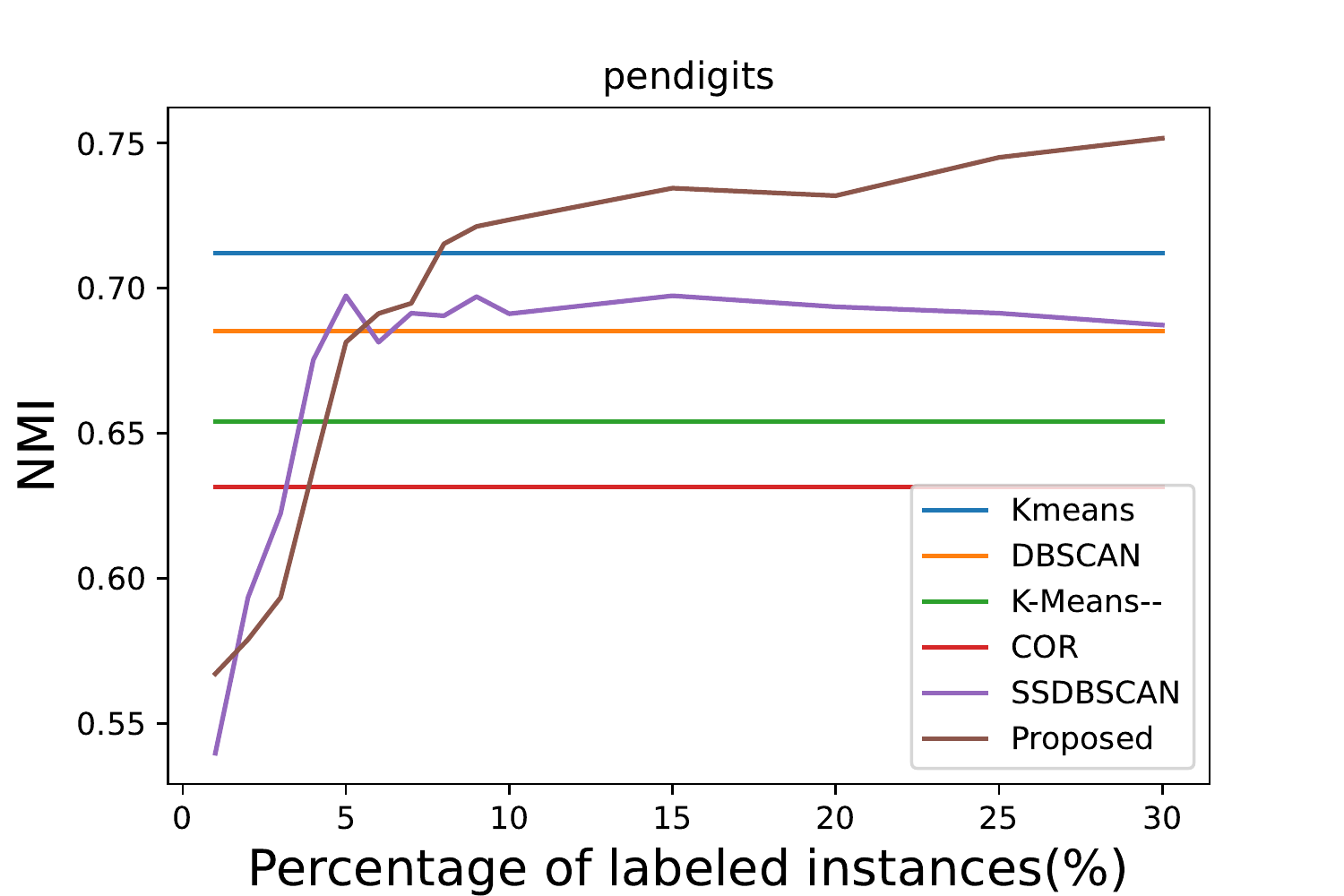}\par

\end{multicols}
\caption{Evaluation Results of the Clustering in NMI}
\label{fig:eval-cluster-nmi}
\end{figure*}

\begin{figure*}
\begin{multicols}{3}
    \includegraphics[width=\linewidth]{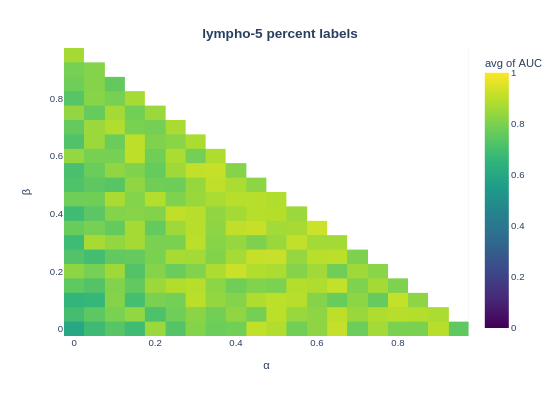}
    \par 
    \includegraphics[width=\linewidth]{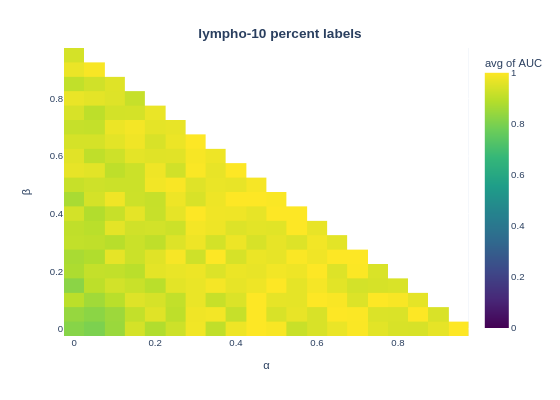}\par 
    \includegraphics[width=\linewidth]{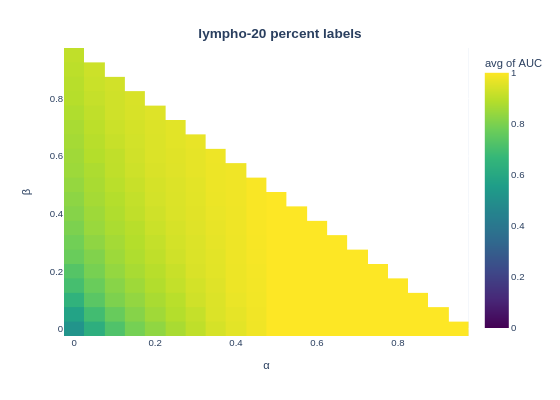}\par 
\end{multicols}
\begin{multicols}{3}
    \includegraphics[width=\linewidth]{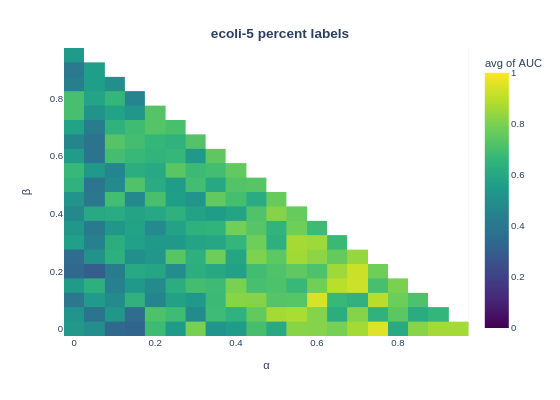}
    \par
    \includegraphics[width=\linewidth]{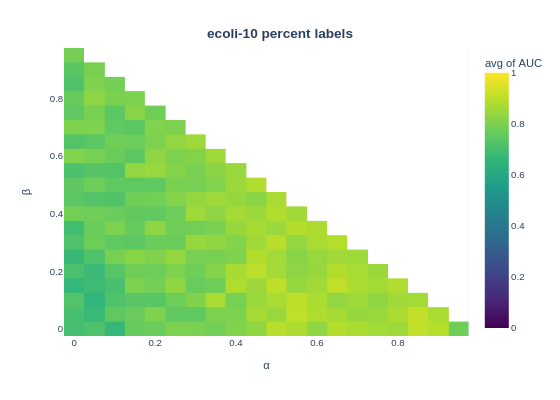}\par 
    \includegraphics[width=\linewidth]{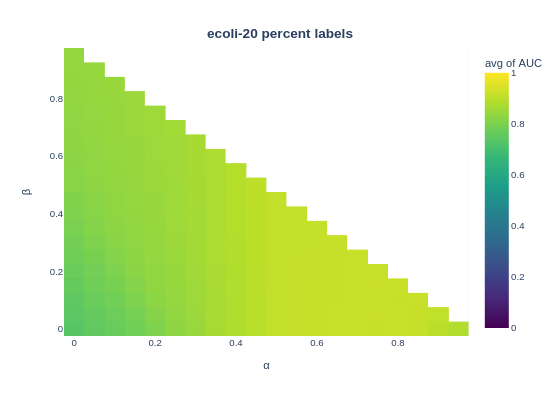}\par 
\end{multicols}

\begin{multicols}{3}
    \includegraphics[width=\linewidth]{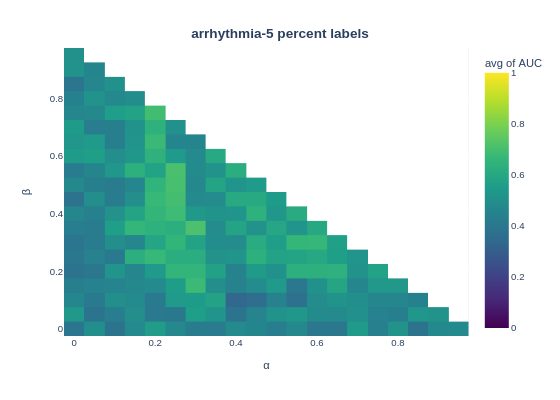}
    \par
    \includegraphics[width=\linewidth]{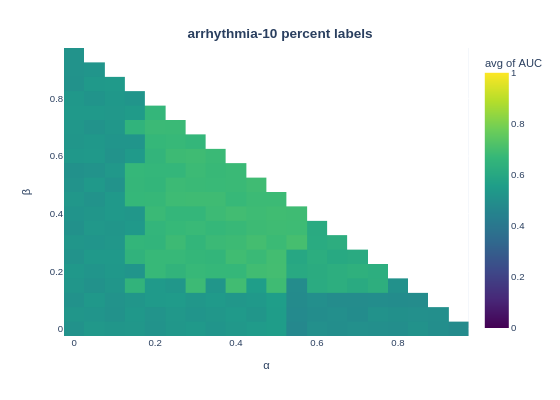}\par 
    \includegraphics[width=\linewidth]{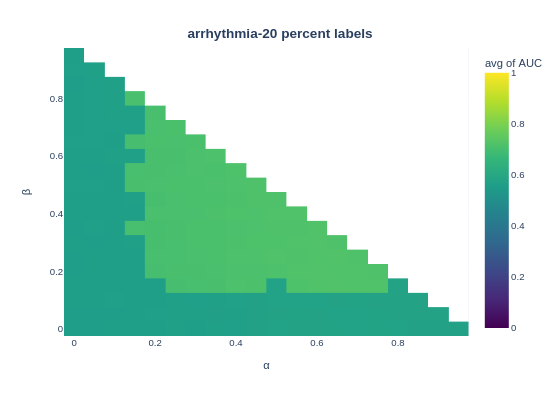}\par 
\end{multicols}

\begin{multicols}{3}
    \includegraphics[width=\linewidth]{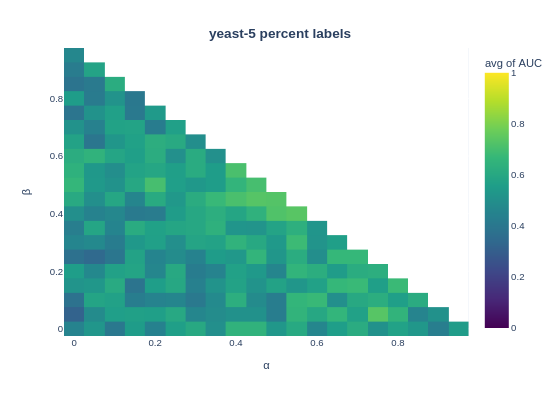}
    \par
    \includegraphics[width=\linewidth]{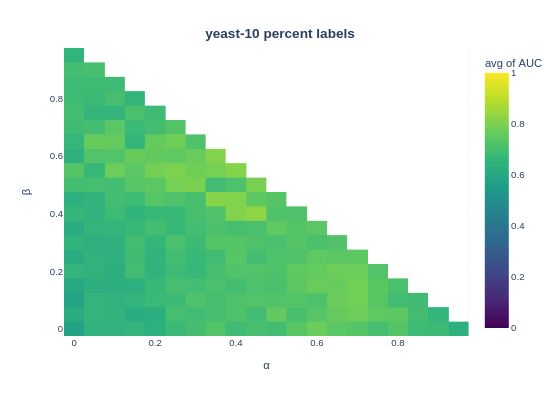}\par 
    \includegraphics[width=\linewidth]{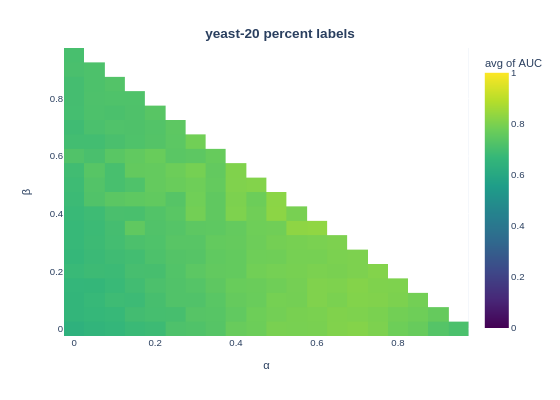}\par 
\end{multicols}
\begin{multicols}{3}
    \includegraphics[width=\linewidth]{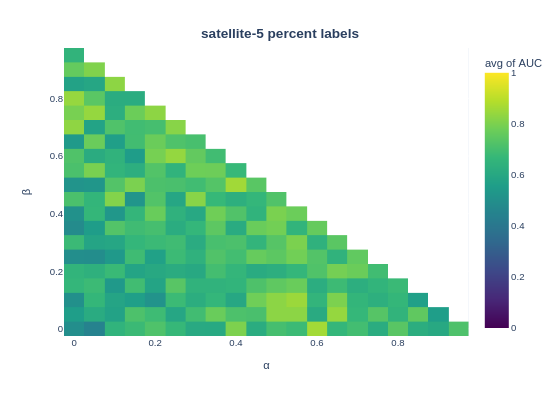}
    \par
    \includegraphics[width=\linewidth]{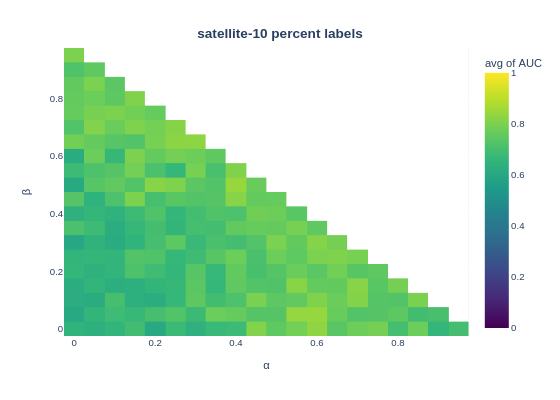}\par 
    \includegraphics[width=\linewidth]{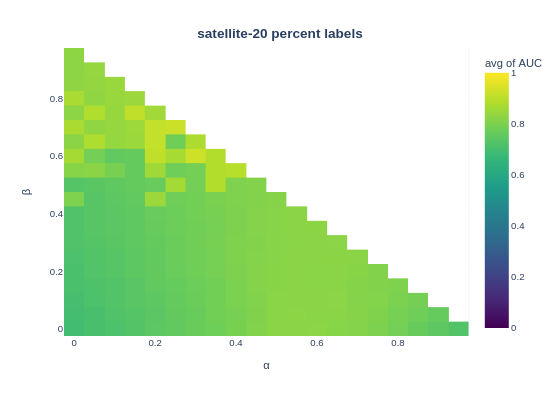}\par 
\end{multicols}

\caption{Evaluation Results of the Parameter Sensitivity for Outlier Detection. From left to right: 5 \%, 10 \% and 20\% labels}
\label{fig:eval-sensitivity}
\end{figure*}

\begin{figure*}
\begin{multicols}{3}
    \includegraphics[width=\linewidth]{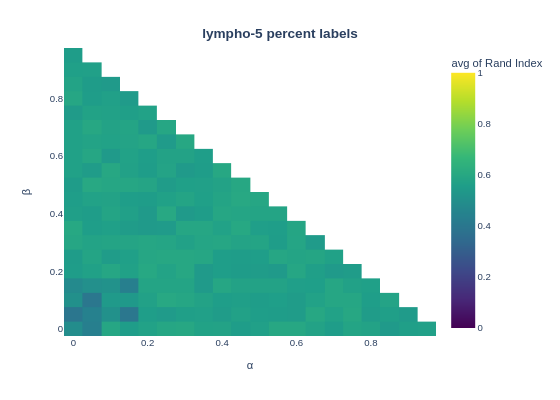}
    \par 
    \includegraphics[width=\linewidth]{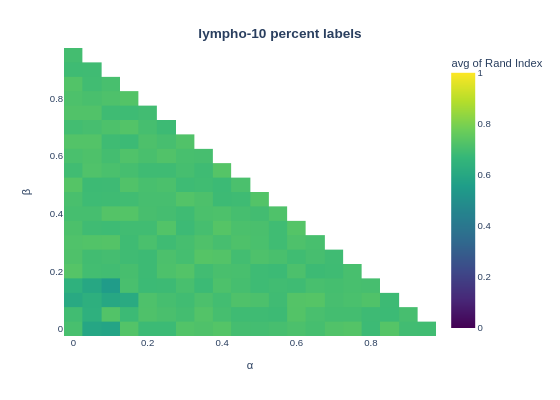}\par 
    \includegraphics[width=\linewidth]{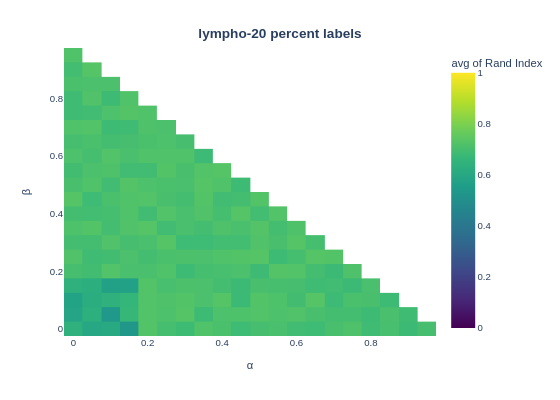}\par 
\end{multicols}
\begin{multicols}{3}
    \includegraphics[width=\linewidth]{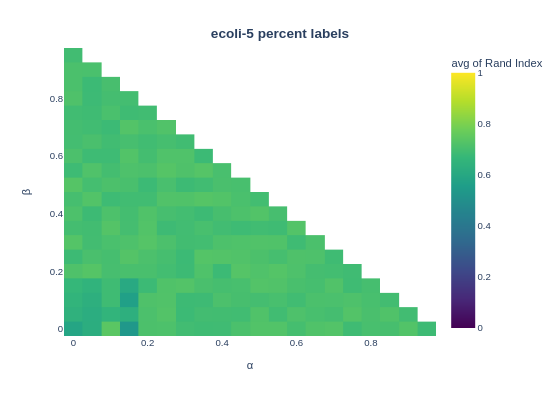}
    \par
    \includegraphics[width=\linewidth]{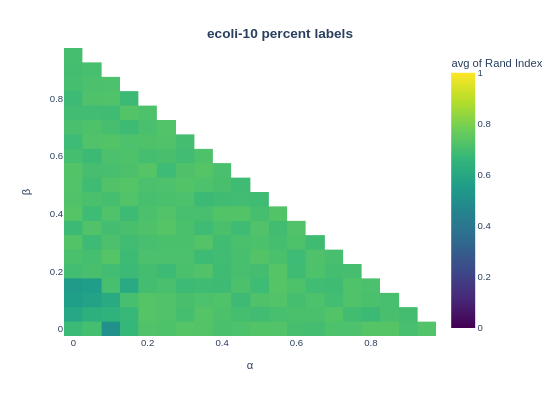}\par 
    \includegraphics[width=\linewidth]{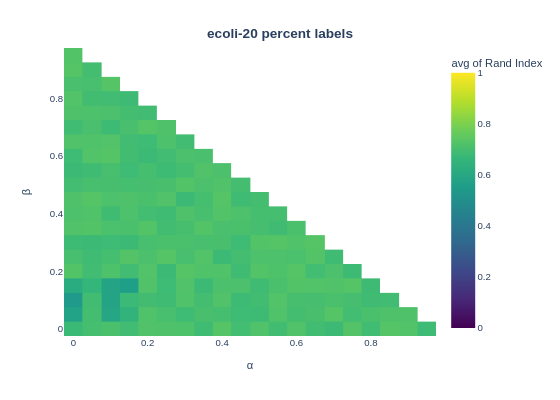}\par 
\end{multicols}

\begin{multicols}{3}
    \includegraphics[width=\linewidth]{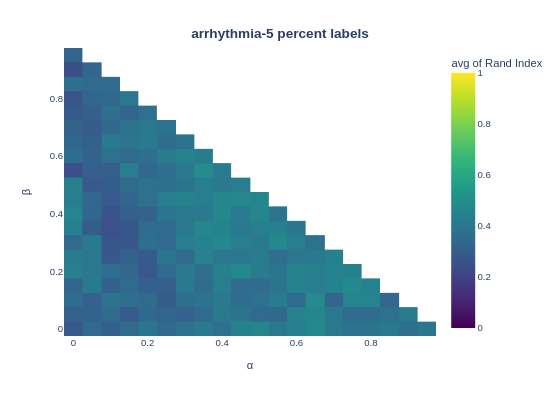}
    \par
    \includegraphics[width=\linewidth]{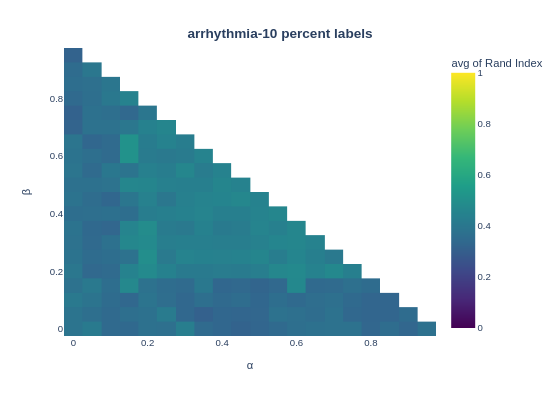}\par 
    \includegraphics[width=\linewidth]{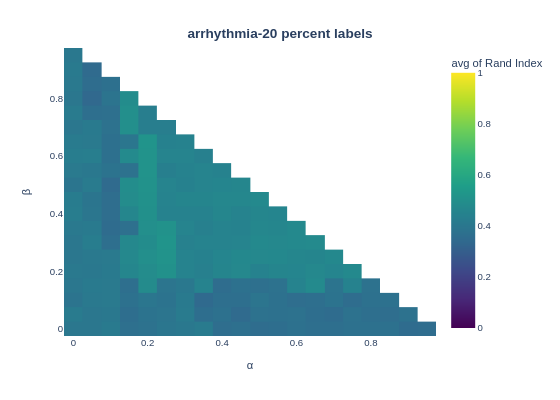}\par 
\end{multicols}

\begin{multicols}{3}
    \includegraphics[width=\linewidth]{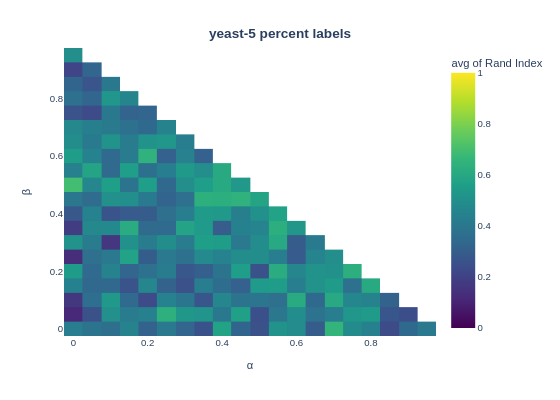}
    \par
    \includegraphics[width=\linewidth]{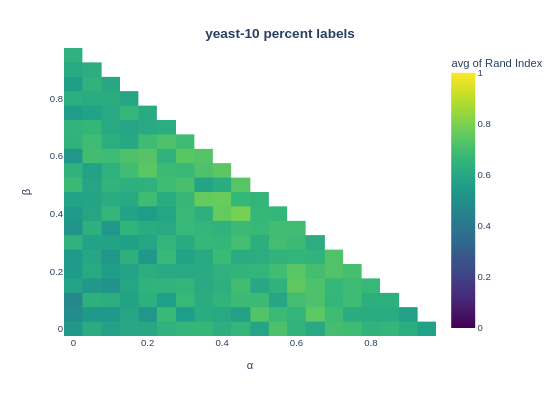}\par 
    \includegraphics[width=\linewidth]{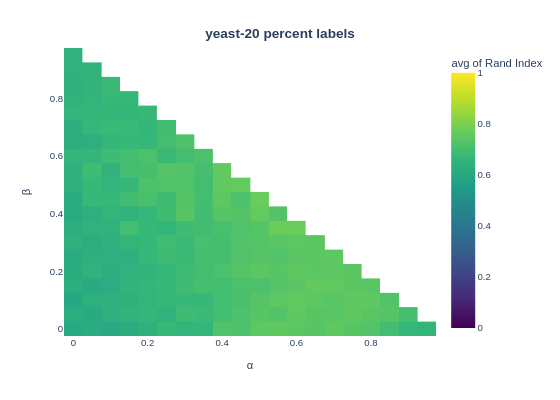}\par 
\end{multicols}
\begin{multicols}{3}
    \includegraphics[width=\linewidth]{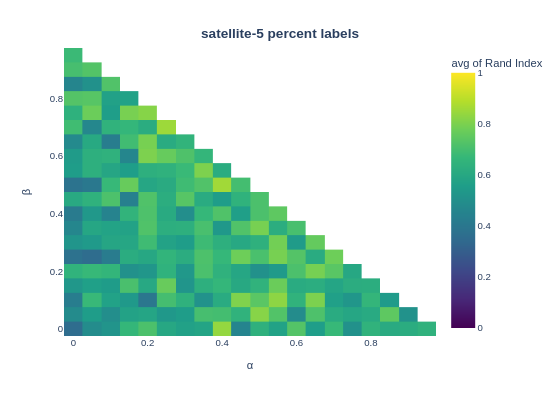}
    \par
    \includegraphics[width=\linewidth]{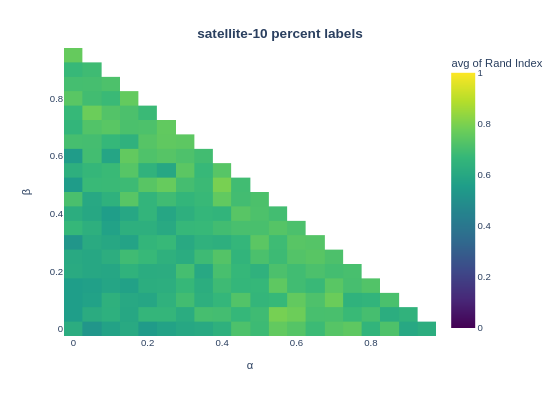}\par 
    \includegraphics[width=\linewidth]{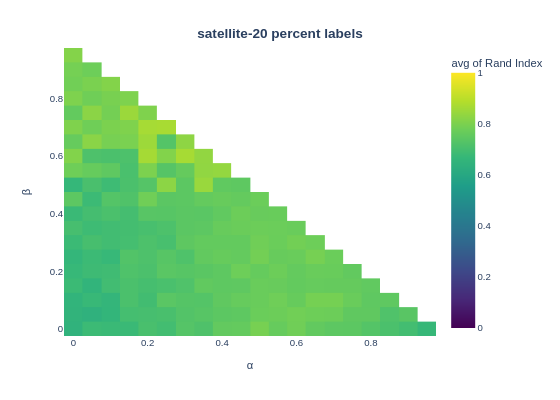}\par 
\end{multicols}

\caption{Evaluation Results of the Parameter Sensitivity for Clustering. From left to right: 5 \%, 10 \% and 20\% labels}
\label{fig:eval-sensitivity-cluster}
\end{figure*}

To understand how our proposed algorithm compares to others, we use the following algorithms as baseline approaches:
\begin{itemize}
    \item \textbf{Local Outlier Factor (LoF)} This is a density-based algorithm proposed by Breunig et al.~\cite{breunig2000lof}, and widely used in many outlier detection tasks. We use this algorithm as a baseline for outlier detection.
    \item \textbf{Isolation Forest (IF)} The isolation forest algorithm proposed by Liu et al.~\cite{liu2008isolation} is one of the best-performing algorithms used in the field of outlier detection. We use this algorithm as a baseline algorithm for outlier detection.
    \item \textbf{One-Class SVM (OSVM)} OSVM is a model-based outlier detection algorithm inspired by the traditional support vector machine algorithm for classification. In our case, we use OSVM with a Radial basis kernel as a baseline algorithm for outlier detection.
    \item \textbf{K-Means, K-Means-- and COR} K-Means-- and COR are algorithms modified from classic K-Means to incorporate the outlier detection mechanism as explained in Section \ref{related}. We use these two algorithms as a baseline for both clustering and outlier detection, while the we use traditional k-means for clustering only.
    \item \textbf{DBSCAN and SSDBSCAN} These two algorithms are both density-based algorithms for clustering. We use them as baseline algorithms for clustering only. The hyper-parameters for SSDBSCAN are consistent with the ones used in the original paper.
\end{itemize}

\subsection{Hyper-Parameter Tuning}\label{subsec:hyper-parameter}
As illustrated in Section \ref{method}, $\alpha$ and $\beta$ are the most important hyper-parameters used for our proposed algorithm. They control the weight assigned to each of the outlier labels used for outlier detection. Since outlier detection influences clustering in our implementation, we use the average of AUC score for outlier detection and the rand index for clustering to tune the parameters. The best set of hyper-parameters is the one that produces the highest average of AUC and rand index across the validation sets. $MinPts$ is set to 3, which is consistent with the original SSDBSCAN.
To improve the accuracy while avoiding the issue of over-fitting, we use cross-validation. 
However, due to the fact that in many cases outliers only account for a small percentage of the entire dataset, we end up with zero outlier folds in some validation sets. In those cases, we use the Rand Index alone.

\subsection{Empirical Results}\label{subsec:empirical}
Since our proposed approach requires user labels, we randomly set a certain percentage of data in each of the datasets in Table \ref{table:data} as labeled instances. 
To reduce the fluctuation due to the randomization, we run 50 trials for each dataset and report the average results. We tune the hyper-parameters under each trial.
Figure \ref{fig:eval-out} shows the results for outlier detection. Figures \ref{fig:eval-cluster} and \ref{fig:eval-cluster-nmi} are results for evaluation measured in Rand Index and NMI respectively. For both outlier detection and clustering, the x-axis represents the proportion of the dataset that is labeled, illustrated in percentage. The y-axes are the AUC score, Rand Index, and NMI respectively.
As illustrated in Figure \ref{fig:eval-out}, in most cases, our proposed algorithm shows a higher AUC score compared to the baseline approaches especially where the labeled instances account for greater than ten percent.  However, for the case of pendigits, the proposed algorithm does not show superiority in terms of AUC due to the fact that most baseline approaches already perform well. However, the arrhythmia and satellite datasets demonstrate how our proposed algorithm can outperform the baseline approach by a significant margin. A similar trend applies to the clustering case as illustrated in Figure \ref{fig:eval-cluster} and Figure \ref{fig:eval-cluster-nmi}.

\subsection{Analysis of Parameter Sensitivity}
As discussed in Section \ref{method}, $\alpha$ and $\beta$ are two of the major hyper-parameters used to compute the $tScore$ which is then used to calculate the outlierness of each data point. Change in those parameters will directly affect the overall quality of the outlier detection. To measure the sensitivity, we run experiments using the same datasets as shown in Table \ref{table:data}. Instead of tuning the parameters per trial, we run multiple parameter sets.
Figure \ref{fig:eval-sensitivity} illustrates our results -- the x-axis and y-axis represent the $\alpha$ and $\beta$ values respectively. The color of each cell represents the mean accuracy measure (AUC for outlier detection and Rand Index for clustering) computed by averaging 50 trials of random labeling. Notice that the accuracy shown in the upper right part of every plot is always equal to 0. This is caused by the constraint that $\alpha + \beta \leq 1$ as discussed in Section \ref{method}. Each row of plots represent the same dataset but with different percentage of labeled instances. From left to right, each plot in the row represents 5\%, 10\% and 20\% of labeled instances. This helps to visualize the change in quality of the results under different percentage of labels for each set of hyper-parameters.

Three findings can be drawn from these parameter sensitivity plots. 
First, for outlier detection, more labels generally result in less sensitivity. This pattern shows in the heatmap plots in the smoothness of the cell colors. For each dataset, more percentage of labels will lead to a more uniform set of colors for all cells and vice versa. In other words, for models trained with fewer labels, a small change in the parameters is likely to result in a significant change in accuracy. 

Second, for the clustering, we see that for datasets where outliers only account for a small percentage of the data, the difference in the percentage of labels does not significantly affect the sensitivity to parameter change.
For example, in Figure \ref{fig:eval-sensitivity-cluster}, the lympho and ecoli datasets, which have few outliers, show little variation. 
This lack of sensitivity is caused by the fact $lScore$ and $simScore$ are not directly used in the clustering process. However, due to the fact that they affect the $tScore$ and thus affect the labeling of $R_{O}$, they will still cause some fluctuations in parameter sensitivity. For other datasets where outliers account for a significant amount, parameter sensitivity shares a similar pattern with the outlier detection process: more labeling leads to less change in sensitivity.
From the plots, we can see they share a similar optimal region (region with high quality) with the outlier detection.

Third, in most cases for both outlier detection and clustering, if we set both  $\alpha$ and $\beta$ values low, it generally results in low quality (AUC for outlier detection and rand index for clustering).
The bottom left corner of the heatmap is usually relatively low in quality compared to other parts. In other words, it is not recommended to put an extremely high weight on the outlier detection's $SimScore$ parameter.

\section{Conclusion}
In this paper, we have proposed a semi-supervised density-based algorithm that utilizes user labels to combine outlier detection with clustering. The algorithm consists of three steps: firstly, it runs the regular SSDBSCAN algorithms and computes the three scores: reachability-score, local-density-score and similarity score. Each score is a measure of the outlierness from different perspective. Secondly, it computes the weights for reliable outliers and instances of normal instsances based on the three scores. Finally, a classifier of user's choice is run with the instance weights. As for the evaluation, we have run our algorithm against some of the state-of-art clustering and outlier detection algorithms and the final results are listed separately for outlier detection and clustering.  The results indicate that our algorithm can achieve superior results on datasets with only a relatively small percentage of labels. \par
In the future, we plan to address the major limitation of our approach that currently, we assume the labels are direct user assignment of the instances to the clusters. However, sometimes we should only expect users can provide very weak insight such as cannot-line or must-link constraints between few instances instead of actual cluster assignment.

\ifCLASSOPTIONcompsoc
  \section*{Acknowledgments}
  This work is supported by DePaul University.
\else
  \section*{Acknowledgment}
\fi

 


\ifCLASSOPTIONcaptionsoff
  \newpage
\fi


\begin{IEEEbiography}[{
\includegraphics[width=1.09in,height=1.8in,keepaspectratio]{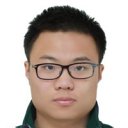}}]
{Jiahao Deng}
Jiahao Deng received the B.S. degree in Business Analytics from the University of Iowa in the USA, in 2017, and the M.S. degree in Computer Science with an emphasis in Data Science from DePaul University, Chicago, IL, USA, in 2019. He is also working toward the Ph.D. degree in Computer Science with DePaul University. He is currently working as a Graduate Research Assistant with the Laboratory for Interactive Human-Computer Analytics. His research interests include the application of data science techniques in data visualization, Robotics, and Computer Vision.
\end{IEEEbiography}

\begin{IEEEbiography}[{
\includegraphics[width=1.09in,height=1.8in,keepaspectratio]{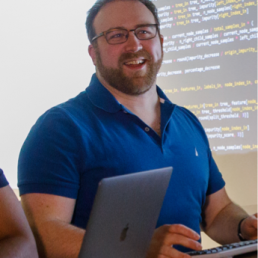}}]
{Eli T. Brown}
Eli T. Brown is an Associate Professor in the College of Computing and Digital Media (CDM) at DePaul University. He earned his B.A. from Cornell University in Computer Science and Math, and his Ph.D. and M.S. in Computer Science from Tufts University. His research revolves around integrating data visualization and machine learning for more effective data analytics and decision making, focusing on how human and machine can collaborate on complex tasks that neither can complete alone. He directs the Laboratory for Interactive Human-Computer Analytics (LIHCA; lihca.io), which develops new interactive machine learning technology and pursues applications to  address the needs of collaborators in a variety of fields including biomedical, biotechnology and journalism.  
\end{IEEEbiography}

\end{document}